\documentclass[journal]{IEEEtran}
\ifCLASSINFOpdf
\else
\fi
\usepackage{amsmath,amssymb,amsfonts}
\usepackage{amsthm}
\usepackage{algorithm}
\usepackage{algorithmic}
\usepackage{amsfonts,amssymb}
\usepackage{graphicx}
\usepackage{subfigure}
\usepackage{booktabs}
\usepackage{color}
\usepackage{xcolor}
\usepackage{soul}
\usepackage{multirow}
\usepackage{boldline} 

\begin{document}
%
\title{Nonnegative-Constrained Joint Collaborative Representation with Union Dictionary for Hyperspectral Anomaly Detection}
%
%
\author{Shizhen Chang,~\IEEEmembership{Member,~IEEE,}
Pedram Ghamisi,~\IEEEmembership{Senior Member,~IEEE}
\thanks{Shizhen Chang is with the Institute of Advanced Research in Artificial Intelligence (IARAI), 1030 Vienna, Austria (e-mail: shizhen.chang@iarai.ac.at).}
\thanks{Pedram Ghamisi is with the Institute of Advanced Research in Artificial Intelligence (IARAI), 1030 Vienna, Austria; Helmholtz-Zentrum Dresden-Rossendorf, Helmholtz Institute Freiberg for Resource Technology, Machine Learning Group, 09599 Freiberg, Germany (e-mail: pedram.ghamisi@iarai.ac.at).}}

\markboth{IEEE TRANSACTIONS ON GEOSCIENCE AND REMOTE SENSING,~Vol.~**, No.~**, June~2022}%
{CHANG AND GHAMISI: Nonnegative-Constrained Joint Collaborative
Representation with Union Dictionary for Hyperspectral Anomaly Detection}

\maketitle

\begin{abstract}
Recently, many collaborative representation-based (CR) algorithms have been proposed for hyperspectral anomaly detection. CR-based detectors approximate the image by a linear combination of background dictionaries and the coefficient matrix, and derive the detection map by utilizing recovery residuals. However, these CR-based detectors are often established on the premise of precise background features and strong image representation, which are very difficult to obtain. In addition, pursuing the coefficient matrix reinforced by the general $l_2$-min is very time consuming. To address these issues, a nonnegative-constrained joint collaborative representation model is proposed in this paper for the hyperspectral anomaly detection task. To extract reliable samples, a union dictionary consisting of background and anomaly sub-dictionaries is designed, where the background sub-dictionary is obtained at the superpixel level and the anomaly sub-dictionary is extracted by the pre-detection process. And the coefficient matrix is jointly optimized by the Frobenius norm regularization with a nonnegative constraint and a sum-to-one constraint. After the optimization process, the abnormal information is finally derived by calculating the residuals that exclude the assumed background information. To conduct comparable experiments, the proposed nonnegative-constrained joint collaborative representation (NJCR) model and its kernel version (KNJCR) are tested in four HSI datasets and achieve superior results compared with other state-of-the-art detectors. The codes of the proposed method will be available online\footnote{https://github.com/ShizhenChang/NJCR}.
\end{abstract}


\IEEEpeerreviewmaketitle

\section{Introduction}
\IEEEPARstart{U}{sing} very rich spectral information existing in hyperspectral images (HSIs), we can diagnose the distribution of land covers and recognize specific objects in a scene \cite{7882742, 9186822, WAMBUGU2021102603}. Target detection, which detects targets of interest utilizing the spectral differences between targets and backgrounds, is one of the important applications for hyperspectral image processing. Technically, target detection can be viewed as a special binary classification problem, which identifies the test pixel as a target or background under the binary hypothesis theory. This technique has been applied in military, civil, and other fields to detect, identify, and monitor specific objects \cite{matteoli2010tutorial, 9532003}.

Depending on the availability of prior target information, target detection can be divided into two categories, supervised \cite{zou2017random} and unsupervised \cite{fowler2011anomaly}. The accuracy of supervised target detection methods is highly dependent on that of the target spectra, which are practically hard to obtain \cite{nasrabadi2013hyperspectral}. Unsupervised target detection, which is also referred to as anomaly detection (AD) or outlier detection, experienced rapid development in the past 20 years \cite{stein2002anomaly}.

So far, most classic anomaly detection algorithms focus on constructing a profile of the distribution of backgrounds to identify those objects that do not belong to the profile as anomalies, or designing a statistical or geometric measurement to separate the anomalies from background instances. An example of the latter is the Reed-Xiaoli (RX) detector \cite{reed1990adaptive}, which solves the anomaly detection task through the Mahalanobis distance and the multivariate Gaussian distribution assumption. Two types of the RX have been widely applied: global RX which considers the entire image as the background statistics, and local RX, which estimates the background using local statistics \cite{borghys2011hyperspectral}. In addition to the traditional RX model, several improved methods based on Mahalanobis distance have also been proposed. A blocked adaptive computationally efficient outlier nominator (BACON) was proposed in \cite{billor2000bacon} to iteratively construct the robust background subset, a random selection based anomaly detector (RSAD) was described in \cite{du2010random}. Inspired by non-negative matrix factorization (NNMF), a robust iterative consensus anomaly RX detector is proposed in \cite{amiel2020consensus} which generates clusters for RX tests and uses a weighted consensus voting process to detect anomalies. Besides the Gaussian distributions models, some non-Gaussian detectors were proposed to model the backgrounds, such as the anisotropic super-Gaussian (AS) \cite{adler2009improved} and Elliptically Contoured (EC) t-distributions \cite{veracini2010spectral}. Based on the Gaussian-Markov random field, GMRF \cite{schweizer2000hyperspectral} models the clutter as spatially–spectrally correlated random fields and derives the simplified detection output. Considering the possible non-validation of the original statistical distribution for real-world HSIs, nonlinear version detectors that map the original data space into a high-dimensional feature space to produce a nearly Gaussian distribution were subsequently formulated. Representative algorithms include the kernel RX (KRX) \cite{kwon2005kernel}, support vector data description (SVDD) \cite{banerjee2006support}, robust nonlinear anomaly detector (RNAD) \cite{zhao2014robust}, and selective kernel principal component analysis (KPCA) algorithm \cite{scholkopf1997kernel}, among others. 

Except for assuming the conditional probability density functions (pdfs) or estimating the covariance matrix of the image scene, some nonparametric detectors have also been proposed \cite{matteoli2013background, arisoy2021nonparametric}. Recently, the representation-based methods have been used more frequently. Considering that the necessary prior knowledge of the anomaly dictionary is unknown for anomaly detection, R. Zhao et al. proposed a background joint sparse representation detector \cite{zhao2017hyperspectral}, which learns and updates the weights of the potential background dictionary, and estimates the reconstructed residual of the background. To balance the influence of the dictionaries, collaborative representation-based detectors were proposed by applying the $\ell_2$-norm regularization term to the original sparse representation model. The most typical algorithm of this type is the collaborative representation-based detector (CRD) \cite{li2014collaborative}, which applies the dual-window strategy and a distance-weighted matrix to balance the effect of background dictionaries. Inspired by the local summation strategy, K. Tan et al. proposed a detector based on the collaborative representation and inverse distance weight for hyperspectral anomaly detection \cite{rs11111318}.

Recently, low-rank matrix decomposition has emerged as a powerful tool for image analysis and has been used for anomaly detection by exploiting the sparse matrix. Since the background features are always dense and overlapped, they can be obtained from the lowest rank representation of the HSI pixels, and the relatively rare anomalies can be obtained by computing the residual of the original image and the recovered background. A low-rank and sparse representation (LRASR) method was proposed in \cite{xu2015anomaly}, which designs a dictionary construction strategy for the sparse component. In \cite{zhang2015low}, LSMAD was proposed which employs the Mahalanobis distance for similarity measurement. A low-rank and sparse decomposition with a mixture of Gaussian (LSDM-MoG) was investigated in \cite{li2020low} for a variety of anomalies and noise. A low-rank-based detector was proposed in \cite{cheng2019graph} with the graph and total variation regularization. Additionally, by combining the low-rank and collaborative representation theory, H. Su et al. designed a new anomaly detection model that decomposes the image into background and anomaly components, and extracts anomalies by adding column sparsity constraints \cite{su2020low}.

However, there is still a common defect in these anomaly detection algorithms: the distinction between the background class and the anomaly class needs to be optimized. This problem is limited by several factors, such as the lack of precise features, insignificant differences after image reconstruction, and inappropriate representation models. The challenges of strengthening the suppression of backgrounds and enhancing the separation of the two classes still need to be explored. At present, the dictionary contribution rules of the representation-based methods as well as the low-rank matrix decomposition-based methods either apply a dual concentric window or are trained by dictionary learning methods. However, these processes may not take full account of global information and result in poor time consumption. Additionally, the coefficients solved by partial derivation of the CR model usually contain negative elements that are contrary to the reality of spectral mixing of HSIs.

In this paper, a novel collaborative representation model with a nonnegative constraint and joint learning is proposed for hyperspectral anomaly detection. All coefficients corresponding to the image are jointly derived through the objective function, which also assumes that the coefficients are nonnegative and obey the sum-to-one rule. For better signal recovery, a global union dictionary containing a background part and an anomaly part are utilized for optimization. The background sub-dictionary is learned by calculating the density peaks within superpixels of the image and the anomaly sub-dictionary is constructed by a number of pixels that have larger outputs after the RX detection. The main contributions of the proposed framework can be summarized as follows:
\begin{itemize}
  \item[1.] The traditional representation-based anomaly detection methods solve the objective function pixel-wise \cite{Su2018Hyperspectral,ZGY2020Hyperpsectral}. With a regularization term of the coefficient vector, the error between the test pixel and its reconstructing vector is minimized. The proposed model substitutes the original $\ell_2$-norm with the Frobenius norm and jointly optimizes the whole coefficient matrix of the image. Thus, possible local anomalies and redundant optimization processes can be prevented \cite{6693730}.
   \item[2.] Unlike previous detectors that only assume the background dictionary by means of a dual-concentric window or other feature selection methods \cite{qu2018hyperspectral, 9133150}, our model designs a unified dictionary constructed by reliable backgrounds and potential anomalies. By excluding the anomaly information of the final residual, the unified dictionary helps to better separate the binary classes.
  \item[3.] With the nonnegative and sum-to-one constraints, the proposed model is more consistent with the spectral mixture characteristics \cite{6200362}, and the reconstruction accuracy of the coefficient matrix is improved.

\end{itemize}

The remainder of this paper is organized as follows. Section II gives a detailed introduction to the proposed non-negative constrained joint collaborative representation (NJCR) and kernel NJCR models. The experimental results of the proposed method with other traditional anomaly detectors are presented in section IV. Finally, the conclusions are given in section V.

\begin{figure*}[t]
\centering 
\includegraphics[width=\linewidth]{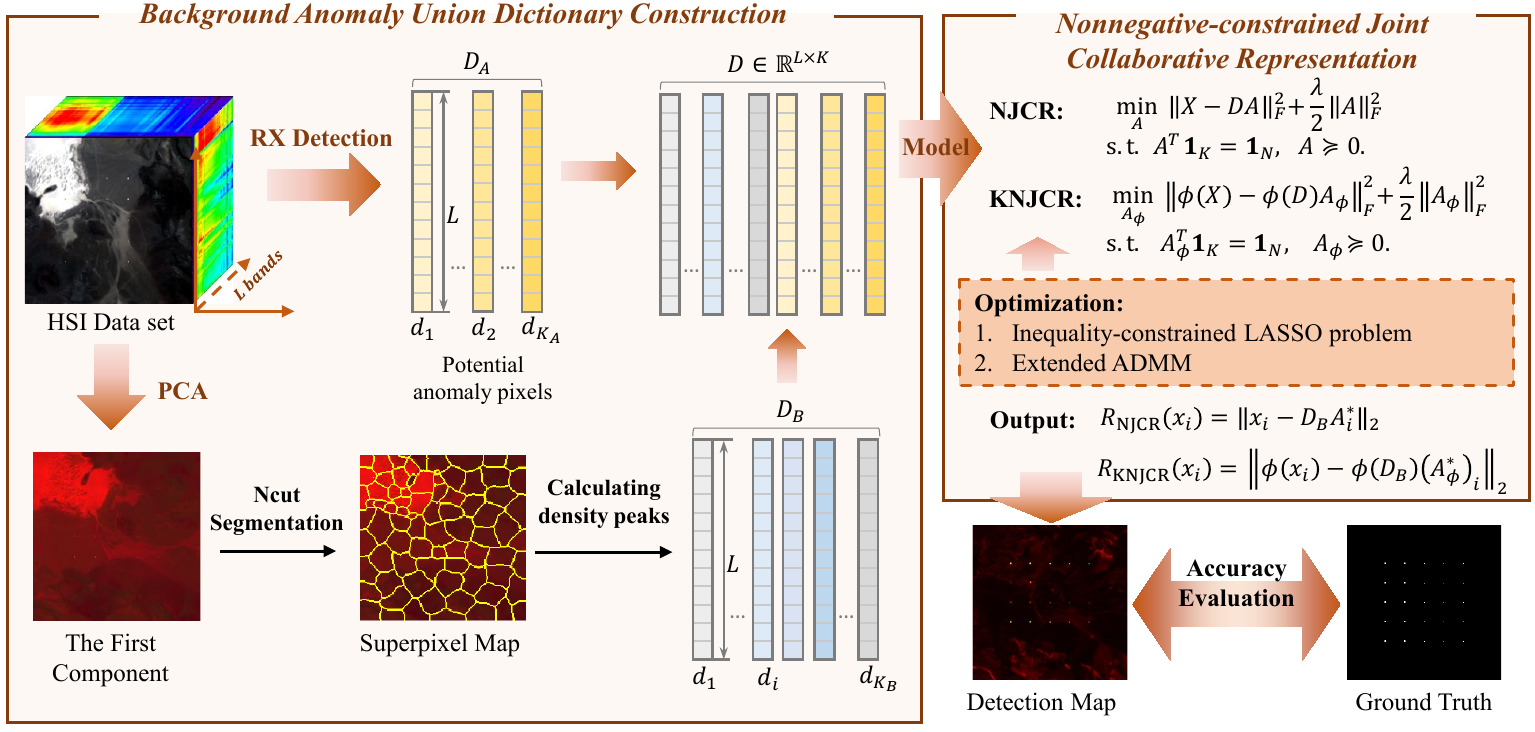}
\caption{The flowchart of the proposed method. A unified background and anomaly dictionary is first constructed, where the background sub-dictionary is obtained by calculating the density peaks of each superpixel and the anomaly sub-dictionary is selected by the RX pre-detection process. Then, the union dictionary is utilized to train the nonnegative-constrained joint collaborative representation models. Finally, the detection result is derived and the accuracies of the proposed models are evaluated.}
\label{flowchart}
\end{figure*}

\section{Nonnegative Constrained Collaborative Representation-based Detection Framework}
In this section, the proposed NJCR model and its kernel version are explored. We will first give an introduction to the general collaborative representation model. Then, a detailed discussion of the construction of the union dictionary, the proposal for the nonnegative-constrained joint collaborative representation (NJCR) model and its kernel version (KNJCR), and the derivation of the models will be described. Fig. \ref{flowchart} gives the overall flowchart of our method.
\subsection{General CR model}
Given a hyperspectral image denoted by $X=[x_1,x_2,\cdots,x_N]\in \mathbb{R}^{L\times N}$, where $L$ is the number of bands and $N$ is the total number of pixels, general collaborative representation-based classifiers (CRCs) \cite{wang2013kernel} approximate each test pixel by the following $l_2$-minimization optimization problem with the training dictionary $D$ from all classes:
\begin{equation}\label{eq:cr}
\hat{\alpha}=\mathrm{arg}\min\limits_{\alpha} \|x-D\alpha\|^2_2+\lambda\|\alpha\|^2_2,
\end{equation}
where $\lambda$ is a regularized scalar. After calculating the partial deviation of $\alpha$ in Eq. (\ref{eq:cr}), and setting the resultant equation to zero, the optimized solution is derived:
\begin{equation}\label{7}
\hat{\alpha}=(D^\mathrm{T}D+\lambda I)^{-1}D^\mathrm{T}x.
\end{equation}
By collaboratively representing the test sample using a training dictionary, the decision rule of the original CRC compares the class-specific representation residual:
\begin{equation}\label{8}
R(x_i)=\|x_i-\hat{x}_i\|_2=\|x_i-D\hat{\alpha}_i\|_2.
\end{equation}
\par The reason CRC works is that it emphasizes the collaborative relationship between training dictionaries with a much lower computational burden compared to sparse representation classifiers (SRCs). In fact, \cite{shi2011face} even argued that it is not the sparse representation but the adoption of collaborative representations in general that play a more crucial role in the success of SRCs.

In this paper, an improved CR model for hyperspectral anomaly detection is proposed based on the following three aspects: 1) Since no prior information is available, we need to define appropriate ways to construct the dictionary matrix, and to expand the recovery residuals of the test samples belonging to the anomaly class and the background class as much as possible. 2) Considering that the coefficient vectors usually have negative elements that violate the non-negative correlation between the dictionary and the test samples, we update the general CR model to better fit the reality. 3) To save the expensive time consuming process of pixel-wise optimization, the joint CR model is designed to minimize the reconstruction errors of the whole image jointly. A detailed description of these three parts will be shown subsequently.
\subsection{The Construction of the Union Dictionary}
For representation-based methods, giving an over-completed training dictionary to accurately approximate the test samples is very important. The supervised target detection research can construct a union dictionary with backgrounds and targets that are usually regarded as prior knowledge \cite{9133150}. But for the proposed anomaly detection problem, neither the backgrounds nor the anomalies are known previously. Considering that the anomaly information is rather difficult to obtain, previous representation-based detectors only learn or select potential background information as the dictionary to estimate the test samples. Thus, the detection result will be influenced if the background is contaminated or not representative enough. To improve the approximation accuracy, we design a union dictionary consisting of a background sub-dictionary and an anomaly sub-dictionary.

To construct the background part of the dictionary, we first use the over-segmentation process to partition the image into superpixels, then we calculate the density peak of each superpixel and select pixels that potentially belong to backgrounds as the background part.

The superpixel is a relatively new concept for image analysis tasks. By incorporating neighboring pixels into the same material of the thematic map, superpixel segmentation algorithms can efficiently reduce the computational complexity, as they can assist in reducing the complexity of images from hundreds of thousands of pixels to only a few hundred superpixels. In hyperspectral imagery, superpixels are often regarded as a uniform parcel of the individual land-cover objects \cite{7097693}. So useful background information can be extracted by the over-segmentation procedure.

Current state-of-the-art graph-based superpixel segmentation approaches include Ren and Malik’s normalized cut (NCut) \cite{shi2000normalized}, Felzenszwalb and Huttenlocher's (FH) superpixel generation \cite{felzenszwalb2004efficient}, and entropy rate (ER) superpixels \cite{liu2011entropy}. The method utilized in this paper is the NCut method that uses mathematical graph theory to connect the spatially contiguous pixels in the image.

Based on the graph theory, a weighted undirected graph $G=(V,E)$ can be constructed from the image, where $V=(v_1,...,v_n)$ is the set of nodes and $E=\{e_{i,j}\}$ represents the connecting edges between the nodes $v_i$ and $v_j$. For a weighted undirected graph, each edge has an associated weight $w_{i,j}$ that shows the similarity between the nodes, and we always have $w_{i,j}=w_{j,i}$. Intuitively, the edges define the ``neighborhood'' relationship between pixels, and the weights show how ``close'' they are \cite{zhang2016biased,gillis2012hyperspectral}. With the over-segmentation process, the NCut method separates $V$ into two or more groups such that nodes that have larger weights are in the same cluster, while nodes that have small weights are divided into different clusters.

If we assume $V$ is partitioned into two subgroups $A$ and $B$, the \textit{cut} between them can be defined as:

\begin{equation}\label{eq1}
\textit{cut}(A,B)=\sum\nolimits_{i\in A,j\in B}w_{i,j}.
\end{equation}

Theoretically, the optimal segmentation exists when Eq. (\ref{eq1}) is maximized. However, it usually results in segmenting the nodes into small groups of outliers that lie far away and the remaining instances. To overcome this problem, the NCut method defines the \textit{association} of $A\subset V$ as:
\begin{equation*}
\textit{assoc}(A)=cut(A,V)=\sum\nolimits_{i\in A,v\in V}w_{i,v}.
\end{equation*}

So the criterion of the normalized cut is defined as:
\begin{equation}\label{eq2}
ncut(A,B)=\frac{cut(A,B)}{\textit{assoc}(A)}+\frac{cut(A,B)}{\textit{assoc}(B)}.
\end{equation}

Detailed descriptions to minimize $ncut(A,B)$ can be found in \cite{shi2000normalized}.

After obtaining the superpixel map with approximate size, we adopt the density peak (DP) clustering method to calculate the density of each pixel in the superpixel and extract those pixels that have larger peaks to create the training samples of the background sub-dictionary. Assume a superpixel $X^S\in\mathbb{R}^{L\times n}$ contains a number of spectral pixels $[x^S_1,x^S_2,\cdots,x^S_{n}]$, where $n$ is the number of pixels located in this superpixel. The Euclidean distance between two pixels $x_i$ and $x_j$ can be expressed as:
\begin{equation}\label{eq4}
d_{ij}=||x_i-x_j||_2.
\end{equation}
To calculate the local density of each pixel, several thresholds can be utilized, such as the cut-off kernel and the Gaussian kernel. It has been proved that the Gaussian kernel threshold can decrease the negative impact of statistical errors caused by using fewer samples \cite{tu2020hyperspectral}. So it is adopted in this paper, and the local density $\gamma_i$ of the pixel $x_i$ can be defined as:
\begin{equation}\label{eq5}
\gamma_i=\sum\nolimits_j\exp(-\frac{d^2_{ij}}{d^2_c}),
\end{equation}
where $d_c$ is the cut-off distance. Then, the minimum distance $\delta_i$ between $x_i$ and other higher-density data points is defined as:
\begin{equation}\label{eq6}
\delta_i=\min_{j:\gamma_j>\gamma_i}d_{ij}.
\end{equation}

The values of $\gamma_i$ and $\delta_i$ can generate a two-dimensional decision
graph used to choose the cluster centers \cite{wang2020mcdpc}. Data points with relatively higher $\gamma$s and $\delta$s are usually viewed as cluster centers and can be selected as the representative training samples of the current superpixel $X^s$. The background sub-dictionary $D_B$ is then constructed by a union of all these training samples.

For the anomaly part, we first implement the RX detection process to the whole image. According to the approximated general likelihood ratio test (GLRT), the detection rule of the test pixel $x_i$ under the RX detector is written as:
\begin{equation}
    R_{RX}(x_i)=(x_i-\mu)^\top C^{-1}(x_i-\mu),
\end{equation}
where $\mu$ and $C$ represent the mean vector and the covariance of the HSI. Then, a number of pixels that have relatively larger outputs are selected to construct the anomaly sub-dictionary $D_A$.

Finally, the dictatory $D$ that has a total of $K$ samples is constructed by the union of the background sub-dictionary $D_B$ and the anomaly sub-dictionary $D_A$:
\begin{equation}\label{eq7}
D=[D_B \ D_A].
\end{equation}

\subsection{Nonnegative-Constrained Joint Collaborative Representation (NJCR) Model and Kerneled NJCR (KNJCR)}
Assuming that the HSI pixels with high spectral similarities can be approximated by the given global union dictionary, these pixels are dominant in the same subspaces and can be represented as:
\begin{equation}\label{eq8}
\begin{split}
X&=[x_1\ x_2\ \cdots \ x_N]\\
&=[D\alpha_1+e_1 \ D\alpha_2+e_2 \ \cdots \ D\alpha_N+e_N] \\
&=D\underbrace{[\alpha_1 \ \alpha_2 \ \cdots \ \alpha_N]}_{A}+E,
\end{split}
\end{equation}
where $A$ is the set of all coefficient vectors corresponding to HSI pixels, and $E$ is the noise matrix. Note that the spectral signals have nonnegative correlation with the dictionary, and the sum of the weights in their dominant subspaces should equal one, which means that the sum of each coefficient vector is 1.

Combining the nonnegative and sum-to-one conditions of the coefficient vectors, the aforementioned problem can be optimized by solving the following nonnegative-constrained joint collaborative representation (NJCR) model:
\begin{equation}\label{eq9}
\begin{split}
\min\limits_A\ & \|X-DA\|^2_F+\frac{\lambda}{2}\|A\|_F^2 \\
\mathrm{s.t.}\ & A^\mathrm{T}\textbf{1}_K=\textbf{1}_N, \ \ A\succeq0.
\end{split}
\end{equation}
where $\textbf{1}_K$ and $\textbf{1}_N$ are all one vectors with sizes of $K$ and $N$, respectively.

To provide a further nonlinear analysis for the NJCR model, its kernel version can be explored. Suppose there exists a nonlinear feature mapping function $\phi(\cdot)$ that maps the HSI data and the dictionary to a kernel induced space: $X\rightarrow\phi(X)$, $A\rightarrow\phi(A)$; the proposed NJCR model can be reformulated as its Gaussian kernel version, which is referred to as the KNJCR:
\begin{equation}\label{eq10}
\begin{split}
\min\limits_{A_\phi}\ & \|\phi(X)-\phi(D)A_\phi\|^2_F+\frac{\lambda}{2}\|A_\phi\|_F^2 \\
\mathrm{s.t.}\ &A_\phi^\mathrm{T}\textbf{1}_K=\textbf{1}_N, \ \ A_\phi\succeq0.
\end{split}
\end{equation}

\subsection{Optimization}
Referring to \cite{xu2019generalized}, which solves the inequality-constrained LASSO problem, we utilize the extended alternating direction method of multipliers (ADMM) with slack variables to solve the proposed optimizers.

\begin{algorithm}[tb]
\caption{Solving the NJCR model by the extended ADMM algorithm}
\begin{algorithmic}[1]
\REQUIRE $X\in\mathbb{R}^{L\times N}$ and $D\in\mathbb{R}^{L\times K}$.
\ENSURE Set $k=0$, Terminate $\gets$ False. Initialize $\omega^0$, $\Delta^0$ and $\eta^0$ to zero.
\WHILE{(Terminate == False)}
\STATE{update $A^{k+1}$ by solving the following linear equation
\begin{equation*}
\begin{split}
A^{k+1}=&(2D^\mathrm{T}D+\lambda I+\rho I+\rho \textbf{1}_K\textbf{1}_K^\mathrm{T})^{-1}(2D^\mathrm{T}X- \\&\rho(\Delta^k-\omega^k-\textbf{1}_K\textbf{1}_N^\mathrm{T}+\textbf{1}_K(\eta^k)^\mathrm{T})),
\end{split}
\end{equation*}}
\STATE{update $\omega^{k+1}$ as $\omega^{k+1}=(A^{k+1}+\Delta^k)_+$,}
\STATE{update $\Delta^{k+1}$ as $\Delta^{k+1}=\Delta^k+A^{k+1}-\omega^{k+1}$,}
\STATE{update $\eta^{k+1}$ as $\eta^{k+1}=\eta^k+A^{(k+1)\mathrm{T}}\textbf{1}_K-\textbf{1}_N$,}
\STATE{$k \gets k+1$}
\IF{($\|r^{k+1}\|_F\leq\epsilon$ and $\|s^{k+1}\|_F\leq\epsilon)$}
\STATE{Terminate $\gets$ True}
\ENDIF
\ENDWHILE\\
\end{algorithmic}
\textbf{Output:} Optimal coefficient matrix $A^*=A^k$.
\label{A1}
\end{algorithm}

To start with, let us introduce a slack variable $\omega=\alpha$, so that the problem Eq. (\ref{eq9}) is equivalently written as:
\begin{equation}\label{eq11}
\begin{split}
\min\limits_A\ & \|X-DA\|^2_F+\frac{\lambda}{2}\|A\|_F^2+I_{{R}_+}(\omega) \\
\mathrm{s.t.}\ & A^\mathrm{T}\textbf{1}_K=\textbf{1}_N \\
& A-\omega=\textbf{0}_{K\times N}.
\end{split}
\end{equation}
where $I_{R_+}$ is the indicator function for the nonnegative reals,
\begin{equation*}
I_{{R}_+}(\omega_{ij})=\left\{
\begin{aligned}
& 0 \ \ \ \omega_{ij}\geq 0 \\
& \infty \ \ \ \text{otherwise}.
\end{aligned}\right.
\end{equation*}

Hence, by introducing the Lagrange multiplier $\Delta\in\mathbb{R}^{K\times N}$ and $\eta\in\mathbb{R}^K$, the augmented Lagrangian function of (\ref{eq9}), or equivalently (\ref{eq11}), is:
\begin{equation}\label{eq12}
\begin{split}
&\mathcal{L}(A,\omega,\Delta,\eta) = \|X-DA\|^2_F+\frac{\lambda}{2}\|A\|_F^2+I_{R_+}(\omega) \\
& \ \ \ +\rho(\eta^\textrm{T}(A^\mathrm{T}\textbf{1}_K-\textbf{1}_N))+\frac{\rho}{2}||A^\mathrm{T}\textbf{1}_K-\textbf{1}_N||_F^2 \\
& \ \ \ +\rho\textrm{tr}(\Delta^\textrm{T}(A-\omega))+\frac{\rho}{2}||A-\omega||_F^2,
\end{split}
\end{equation}
where $\rho>0$ is the step size defined by user.

Given $\omega^{k}$, $\Delta^k$, and $\eta^{k}$ as the current solution of (\ref{eq12}), the updating steps can be calculated as follows \cite{gaines2018algorithms}:
\begin{eqnarray*}
&&A^{k+1}=\mathop{\arg\min}\limits_{A}\|X-DA\|^2_F+\frac{\lambda}{2}\|A\|_F^2 + \\
&&\ \ \ \ \ \frac{\rho}{2}||A^\mathrm{T}\textbf{1}_K-\textbf{1}_N+\eta^k||_F^2+\frac{\rho}{2}||A-\omega^{k}+\Delta^{k}||_F^2,\\
&&\omega^{k+1}=\mathop{\arg\min}\limits_{\omega}I_{R_+}(\omega)+\frac{\rho}{2}||A^{k+1}-\omega+\Delta^{k}||_F^2,\\
&&\Delta^{k+1}=\Delta^k+A^{k+1}-\omega^{k+1},\\
&&\eta^{k+1}=\eta^k+A^{(k+1)\mathrm{T}}\textbf{1}_K-\textbf{1}_N.
\end{eqnarray*}

So the optimization problems for updating $A^{k+1}$, $\omega^{k+1}$ is solvable:
\begin{itemize}
\item The optimization for updating $A^{k+1}$ is a least squares problem with $\ell_2$ penalty terms. Some calculation yields:
\begin{equation*}
\begin{split}
A^{k+1}=&(2D^\mathrm{T}D+\lambda I+\rho I+\rho \textbf{1}_K\textbf{1}_K^\mathrm{T})^{-1}(2D^\mathrm{T}X- \\&\rho(\Delta^k-\omega^k-\textbf{1}_K\textbf{1}_N^\mathrm{T}+\textbf{1}_K(\eta^k)^\mathrm{T})).
\end{split}
\end{equation*}
\item Due to the special structure of function $I_{{R}_+}(\omega)$, the updated $\omega^{k+1}$ can be written as:
\begin{equation*}
\omega^{k+1}=(A^{k+1}+\Delta^k)_+.
\end{equation*}
\end{itemize}

\begin{algorithm}[tb]
\caption{Solving the KNJCR model by the extended ADMM algorithm}
\begin{algorithmic}[1]
\REQUIRE $X\in\mathbb{R}^{L\times N}$, $D\in\mathbb{R}^{L\times K}$, $\text{K}_{DD}$, and $\text{K}_{DX}$.
\ENSURE Set $k=0$, Terminate $\gets$ False. Initialize $\omega_\phi^0$, $\Delta^0$ and $\eta^0$ to zero.
\WHILE{(Terminate == False)}
\STATE{update $A_\phi^{k+1}$ by solving the following linear equation
\begin{equation*}
\begin{split}
A_\phi^{k+1}&=(2\text{K}_{DD}+\lambda I+\rho I+\rho \textbf{1}_K\textbf{1}_K^\mathrm{T})^{-1}(2\text{K}_{DX}- \\&\rho(\Delta^k-\omega_\phi^k-\textbf{1}_K\textbf{1}_N^\mathrm{T}+\textbf{1}_K(\eta^k)^\mathrm{T})).
\end{split}
\end{equation*}}
\STATE{update $\omega_\phi^{k+1}$ as $\omega_\phi^{k+1}=(A_\phi^{k+1}+\Delta^k)_+$,}
\STATE{update $\Delta^{k+1}$ as $\Delta^{k+1}=\Delta^k+A_\phi^{k+1}-\omega_\phi^{k+1}$,}
\STATE{update $\eta^{k+1}$ as $\eta^{k+1}=\eta^k+A_\phi^{(k+1)\mathrm{T}}\textbf{1}_K-\textbf{1}_N$,}
\STATE{$k \gets k+1$}
\IF{($\|r_\phi^{k+1}\|_F\leq\epsilon$ and $\|s_\phi^{k+1}\|_F\leq\epsilon)$}
\STATE{Terminate $\gets$ True}
\ENDIF
\ENDWHILE
\end{algorithmic}
\textbf{Output:} Optimal coefficient matrix $A^*=A^k$.
\label{A2}
\end{algorithm}

The updating process will stop when convergence is achieved. The stopping criteria of this problem are:
\begin{equation*}
||r^{k+1}||_F\leq\epsilon \ \ \
||s^{k+1}||_F\leq\epsilon,
\end{equation*}
where $r^{k+1}$ and $s^{k+1}$ are the primal and dual residuals \cite{boyd2011distributed}, respectively, given by

\begin{equation*}
r^{k+1}=\begin{bmatrix}
\textbf{1}_K^\mathrm{T}\alpha-\textbf{1}_N^\mathrm{T} \\
\alpha^{k+1}-\omega^{k+1}
\end{bmatrix}, \ \ \
s^{k+1}=-\rho(\omega^{k+1}-\omega^k),
\end{equation*}
and $\epsilon$ denotes the error tolerance specified by the user. In practice, the choice of $\epsilon=10^{-4}$ works well.

Similarly, to solve the kerneled version, Eq. (\ref{eq10}) is also rewritten to a Lagrangian dual function:
\begin{equation}\label{13}
\begin{split}
\mathcal{L}(A_\phi,&\omega_\phi,\Delta,\eta)=\\
&\|\phi(X)-\phi(D)A_\phi\|^2_F+\frac{\lambda}{2}\|A_\phi\|_F^2+I_{R_+}(\omega_\phi)\\
& +\rho(\eta^\mathrm{T}(A_\phi^\mathrm{T}\textbf{1}_K-\textbf{1}_N))+\frac{\rho}{2}||A_\phi^\mathrm{T}\mathbf{1}_K-\mathbf{1}_N||_F^2\\
&+\rho\mathrm{tr}(\Delta^\mathrm{T}(A_\phi-\omega_\phi))+\frac{\rho}{2}||A_\phi-\omega_\phi||_F^2,
\end{split}
\end{equation}
and then the extended ADMM method is utilized to optimize this function. It should be noted that at iteration $k$, the coefficient matrix $A_\phi^{k+1}$ is derived with a kernel version
\begin{equation*}
\begin{split}
A_\phi^{k+1}&=\mathop{\arg\min}\limits_{A_\phi}\|\phi(X)-\phi(D)A_\phi\|^2_F+\frac{\lambda}{2}\|A_\phi\|_F^2+\\
&+\frac{\rho}{2}||A_\phi^\mathrm{T}\textbf{1}_K-\textbf{1}_N+\eta^k||_F^2+\frac{\rho}{2}||A_\phi-\omega_\phi^{k}+\Delta^{k}||_F^2,
\end{split}
\end{equation*}
and the calculation yields that
\begin{equation*}
\begin{split}
A_\phi^{k+1}&=(2\text{K}_{DD}+\lambda I+\rho I+\rho \textbf{1}_K\textbf{1}_K^\mathrm{T})^{-1}(2\text{K}_{DX}- \\&\rho(\Delta^k-\omega_\phi^k-\textbf{1}_K\textbf{1}_N^\mathrm{T}+\textbf{1}_K(\eta^k)^\mathrm{T})),
\end{split}
\end{equation*}
where $\text{K}_{DD}=\kappa(D,D)=\left<\phi(D),\phi(D)\right>$ and $\text{K}_{DX}=\kappa(D,X)=\left<\phi(D),\phi(X)\right>$ are the inner products that project the data into the feature space. The kernel function we adopt is the radial basis function (RBF) kernel $\kappa(x_i,x_j)=\mathop{\exp}(-||x_i-x_j||^2/2\sigma^2)$ \cite{chen2012hyperspectral}.

Finally, by calculating the recovery residual of each pixel that is constructed by the background sub-dictionary, the detection results of the proposed NJCR and KNJCR models are derived:
\begin{eqnarray}
&&R_{\text{NJCR}}(x_i)=||x_i-D_BA_i^\star||_2,\\
&&R_{\text{KNJCR}}(x_i)=||\phi(x_i)-\phi(D_B)(A_\phi^\star)_i||_2.
\end{eqnarray}

In summary, Algorithm 1 and Algorithm 2 show the updating process of the proposed joint collaborative representation models under the extended ADMM implementations.

\section{Experiments and Analysis}
To illustrate the effectiveness of the proposed NJCR and KNJCR models, experiments were conducted on four HSI datasets, which include one simulated hyperspectral dataset and three real-world hyperspectral datasets. Seven traditional anomaly detectors are applied as a comparison. The detection results are evaluated via receiver operation characteristic (ROC) curves, the area under the ROC curve (AUC) values caluclated by $((P_D, P_F))$ and $(P_F, \tau)$, and the anomaly-background separability maps. Then, detailed parametric analysis and ablation study of the proposed nonnegative-constrained joint representation models are discussed.
\subsection{Hyperpsectral datasets}
\begin{figure}[htbp]
\centering
\includegraphics[width=0.85\linewidth]{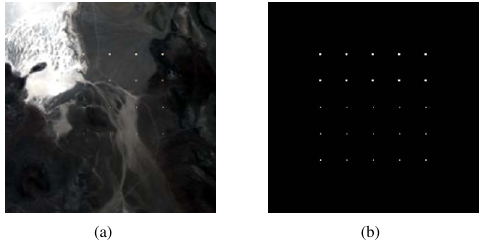}
\centering
\caption{(a) LCVF image scene in true color. (b) Ground truth.}
\label{anno2}
\end{figure}

The first dataset is a simulated dataset collected by the Airborne Visible/Infrared Imaging Spectrometer (AVIRIS) from the Lunar Crater Volcanic Field (LCVF) in Northern Nye County, NV, USA \cite{dong2015maximum,chang2004estimation}. The spectral resolution of this image is 5nm with 224 spectral channels in wavelengths ranging from 370 to 2510nm. The size of the image is $200\times200$ pixels. The original image contains a two-pixel alunite object that is regarded as the target of interest. We collect the spectrum of this alunite object and implant $5\times5$ target panels into the image using a nonlinear mixture model. The size and target spectrum abundance of each panel are shown in Table \ref{Table1}. The true-color image of the LCVF dataset and the corresponding ground truth are shown in Fig. \ref{anno2}.

\begin{table}[tbp]
\caption{Target panels characteristics}
\label{Table1}
\centering
\begin{tabular}{ccc}
\toprule
\hline
Location & Size & Target abundance\\
\hline
$1^{st}$ row & $2\times 2$ pixel & $100\%$, $25\%$, $50\%$, $75\%$, $95\%$\\
\hline
$2^{nd}$ row & $2\times 2$ pixel& $100\%$, $25\%$, $50\%$, $75\%$, $95\%$\\
\hline
$3^{rd}$ row & $1\times 1$ pixel& $100\%$, $25\%$, $50\%$, $75\%$, $95\%$\\
\hline
$4^{th}$ row & $1\times 1$ pixel& $100\%$, $25\%$, $50\%$, $75\%$, $95\%$\\
\hline
$5^{th}$ row & $1\times 1$ pixel& $100\%$, $25\%$, $50\%$, $75\%$, $95\%$\\
\hline
\bottomrule
\end{tabular}
\end{table}

\begin{figure}[htbp]
\centering
\includegraphics[width=0.85\linewidth]{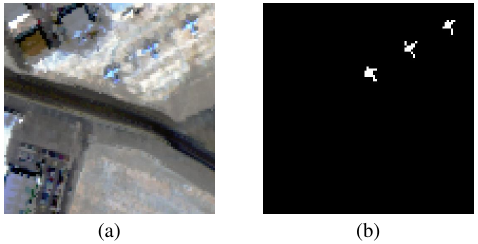}
\centering
\caption{(a) SanDiego image scene in true color. (b) Ground truth.}
\label{anno3}
\end{figure}

The second dataset was captured by the AVIRIS sensor over the area of the SanDiego airport, CA, USA with a spatial resolution of 3.5 m and a spectral resolution of 10 nm. This dataset has 224 spectral bands in total. 189 bands are utilized for the detection task after eliminating the noisy bands. The size of the SanDiego dataset is $100\times100$. In this dataset, three aircrafts, which include 58 pixels, are treated as anomaly targets. The visualized 2-D true-color image and the ground-truth of this dataset are shown in Fig. \ref{anno3}.

\begin{figure}[htbp]
\centering
\includegraphics[width=0.85\linewidth]{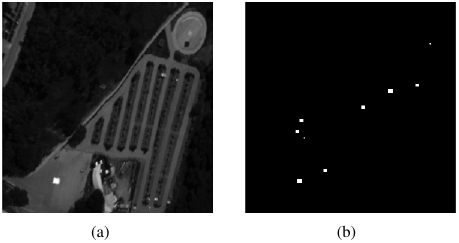}
\centering
\caption{(a) Viareggio image scene in gray color. (b) Ground truth.}
\label{anno4}
\end{figure}

The third dataset is acquired from the airborne Viareggio trial \cite{7430258}, the benchmark hyperspectral detection campaign that took place in Viareggio, Italy in 2013. The sensor collects spectral information ranging from 400 to 1000 nm at a spectral resolution of $1.2$ nm approximately, and the spatial resolution is  about $0.6$ m. A region of $300\times300$ pixels from the original D1F12H2 scene is selected for our experiments. The anomalies to be detected include 191 pixels in total. The image scene and its corresponding ground truth are shown in Fig. \ref{anno4}.

\begin{figure}[tbp]
\centering
\includegraphics[width=0.85\linewidth]{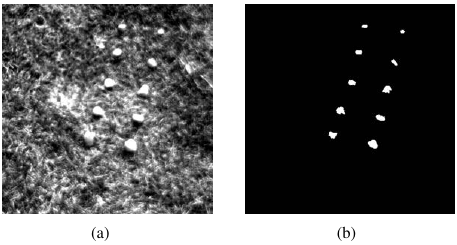}
\centering
\caption{(a) CRi image scene in gray color. (b) Ground truth.}
\label{anno5}
\end{figure}

The fourth hyperspectral dataset was acquired by the CRi Nuance hyperspectral sensor \cite{chang2020subspace}. This image covers an area of $400\times400$ pixels and has 46 spectral channels from 650 to 1100nm. The spectral resolution of this dataset is 10nm. The background land-cover types in this image are various kinds of grasses. There are ten rocks located in the grassy scene, as shown in Fig. \ref{anno5}(a); a total of 1254 pixels were selected to include the anomalies to be detected, as shown in Fig. \ref{anno5}(b).

\subsection{Detection Efficiency}
The detection efficiency of the proposed NJCR and KNJCR models are compared with six representative anomaly detection methods: RX \cite{reed1990adaptive}, CRD \cite{li2014collaborative}, ADLR \cite{qu2018hyperspectral}, LRASR \cite{xu2015anomaly}, PAB-DC \cite{huyan2018hyperspectral}, and RGAE \cite{fan2021hyperspectral} methods, and a kerneled detector: KRX \cite{kwon2005kernel}. To conduct balanced detection performances, all of these algorithms are implemented with optimal parameter settings. The PAB-DC method is conducted in Python on a workstation with Intel Xeon Gold 6254 CPU @ 3.10GHz with 1.58 TB of RAM. Except that, all experiments are conducted in MATLAB on an Intel Core i7-8550U CPU with 16 GB of RAM.

 To construct the union dictionary, the number of samples in the sub-dictionaries should be large enough to represent the most important information of the anomalies and the backgrounds. On the other hand, an appropriate number of samples is also needed to prevent huge time consumption caused by over-segmentation and interference. With reference to the results of the existing publications \cite{xu2015anomaly, zhao2017hyperspectral}, we can find it is reasonable to set the size of background dictionary around $[300, 600]$. By taking the image size and background complexity of the datasets into account, we first segment the image into 100 superpixels and extract 5 representative samples within each superpixel using the density peak clustering method as the background sub-dictionary. And 50 potential anomaly samples are selected after the RX detection as the anomaly sub-dictionary, by experience.

\begin{figure}[tbp]
\centering
\includegraphics[width=0.95\linewidth]{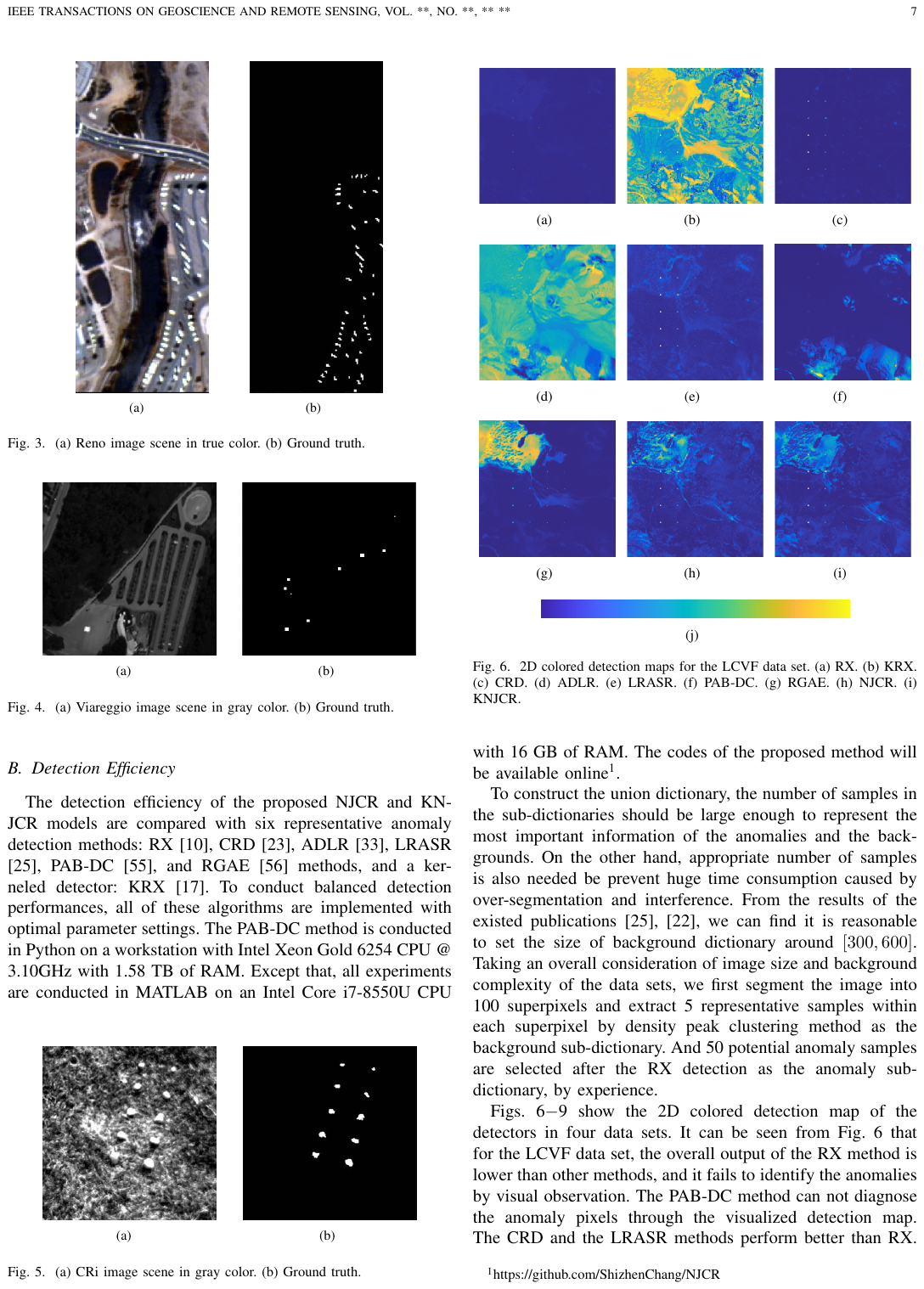}
\caption{2D colored detection maps for the LCVF dataset. (a) RX, (b) KRX, (c) CRD, (d) ADLR, (e) LRASR, (f) PAB-DC, (g) RGAE, (h) NJCR, and (i) KNJCR.}
\label{annotation 6}
\end{figure}

\begin{figure}[htbp]
\centering
\includegraphics[width=0.95\linewidth]{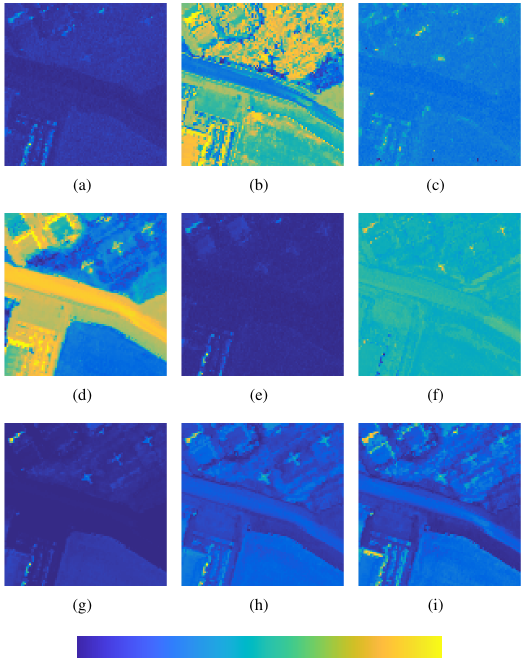}
\caption{2D detection maps for the SanDiego dataset. (a) RX, (b) KRX, (c) CRD, (d) ADLR, (e) LRASR, (f) PAB-DC, (g) RGAE, (h) NJCR, and (i) KNJCR.}
\label{annotation 7}
\end{figure}

\begin{figure}[tbp]
\centering
\includegraphics[width=0.95\linewidth]{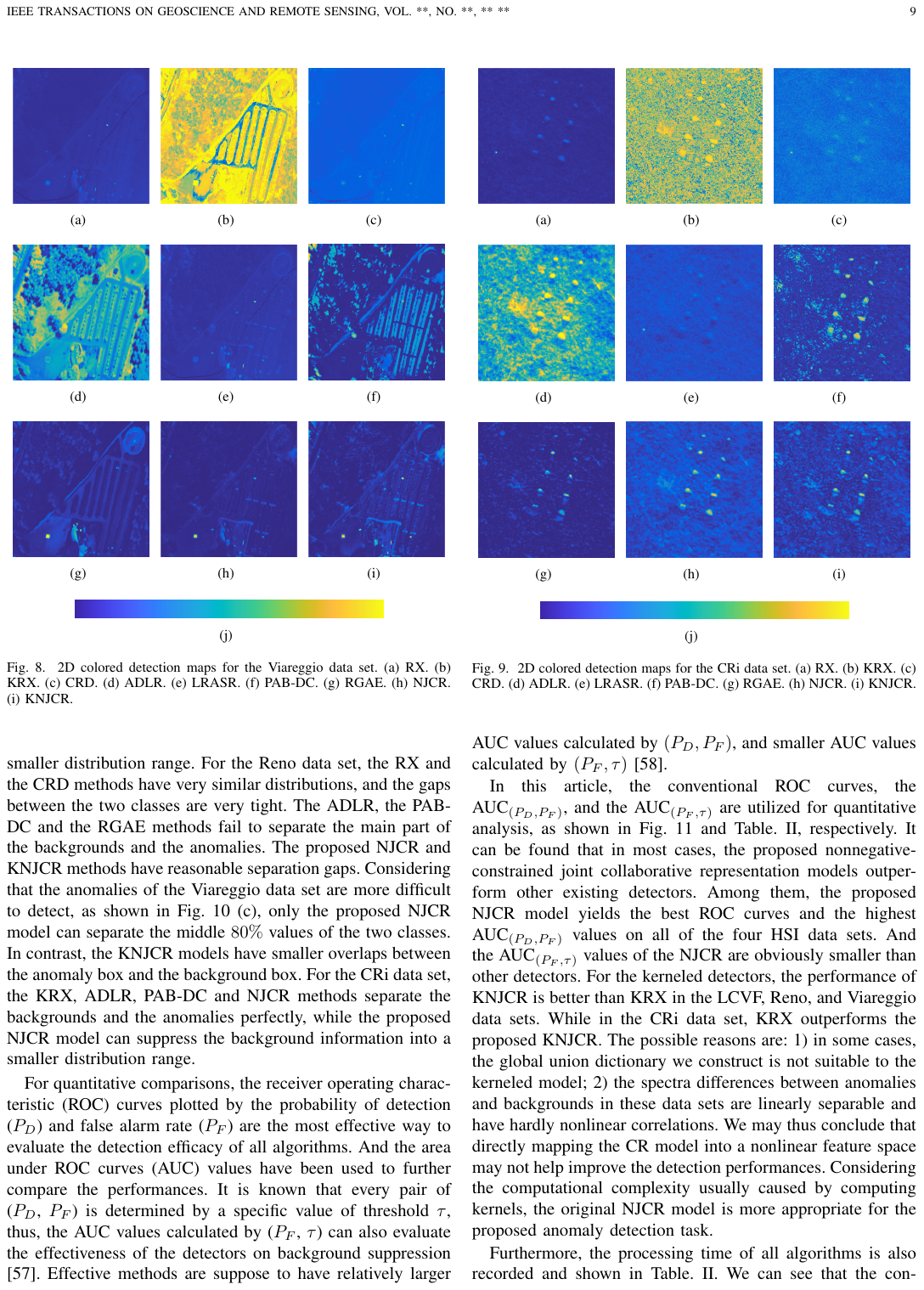}
\caption{2D colored detection maps for the Viareggio dataset. (a) RX, (b) KRX, (c) CRD, (d) ADLR, (e) LRASR, (f) PAB-DC, (g) RGAE, (h) NJCR, and (i) KNJCR.}
\label{annotation 8}
\end{figure}

\begin{figure}[tbp]
\centering
\includegraphics[width=0.95\linewidth]{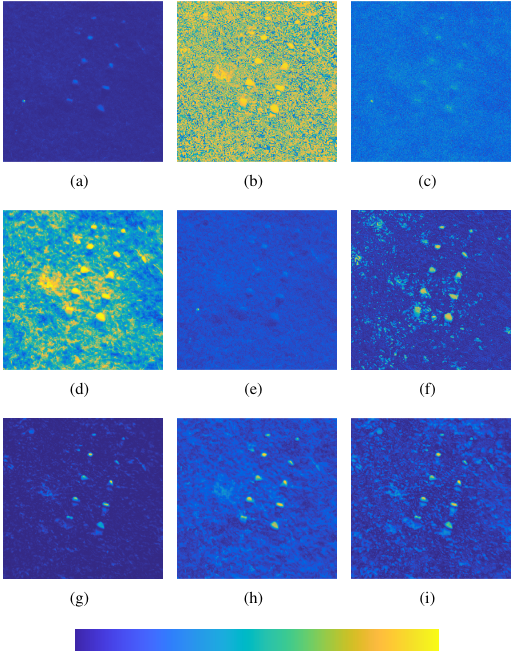}
\caption{2D colored detection maps for the CRi dataset. (a) RX, (b) KRX, (c) CRD, (d) ADLR, (e) LRASR, (f) PAB-DC, (g) RGAE, (h) NJCR, and (i) KNJCR.}
\label{annotation 9}
\end{figure}

Figs. \ref{annotation 6}$-$\ref{annotation 9} show the 2D colored detection map of the detectors in four datasets. It can be seen from Fig. \ref{annotation 6} that for the LCVF dataset, the overall output of the RX method is lower than other methods, and it fails to identify the anomalies by visual observation. PAB-DC can not diagnose the anomaly pixels through the visualized detection map. The CRD and the LRASR methods perform better than RX. KRX, ADLR, and RGAE have higher detection values on both the anomalies and the backgrounds. And comparing with all the other methods, the proposed NJCR and KNJCR models can highlight the anomalies from the background with properly constrained background outputs, although the suppression on the backgrounds is not very strict.

The visual detection maps of all methods on the SanDiego dataset are shown in Fig. \ref{annotation 7}. It is clear that the output values of the KRX and ADLR are much higher than those of the other eight methods, which makes discovering the anomalies very difficult. The visual results of RX, CRD, and RGAE are very similar. Compared to other methods, the proposed NJCR and KNJCR models have balanced visualized outputs between the backgrounds and the anomalies.

Fig. \ref{annotation 8} shows the colored maps of all methods in the Viareggio dataset. The KRX and ADLR methods fail to suppress the output values of backgrounds, which makes it hard to determine the anomaly targets. RX, LRASR, and RGAE have similar outputs as the proposed models, but the anomalies are better highlighted in the proposed models.

\begin{figure}[t]
\centering
\subfigure[]{
\centering
\includegraphics[width=0.6\linewidth]{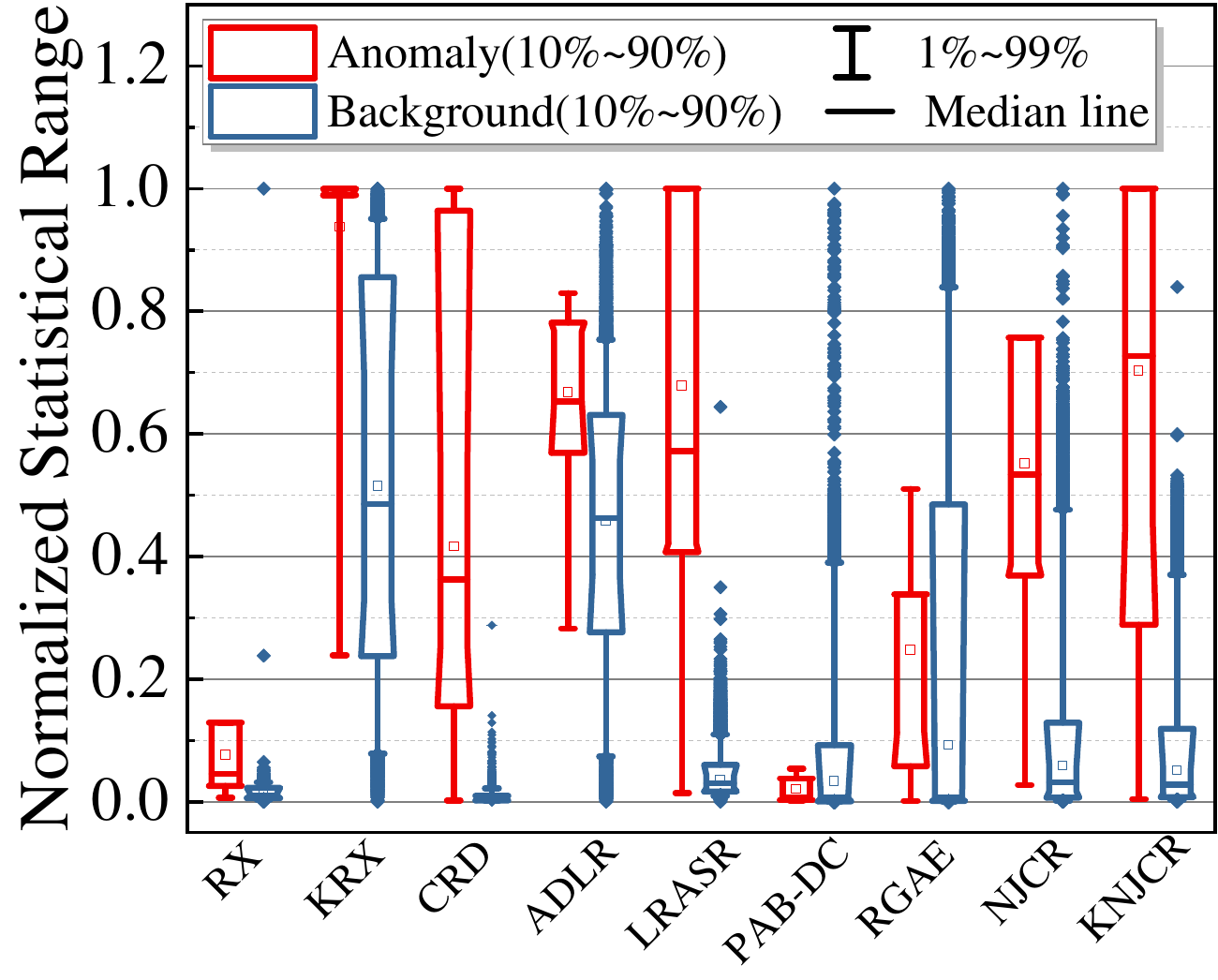}
}

\subfigure[]{
\centering
\includegraphics[width=0.6\linewidth]{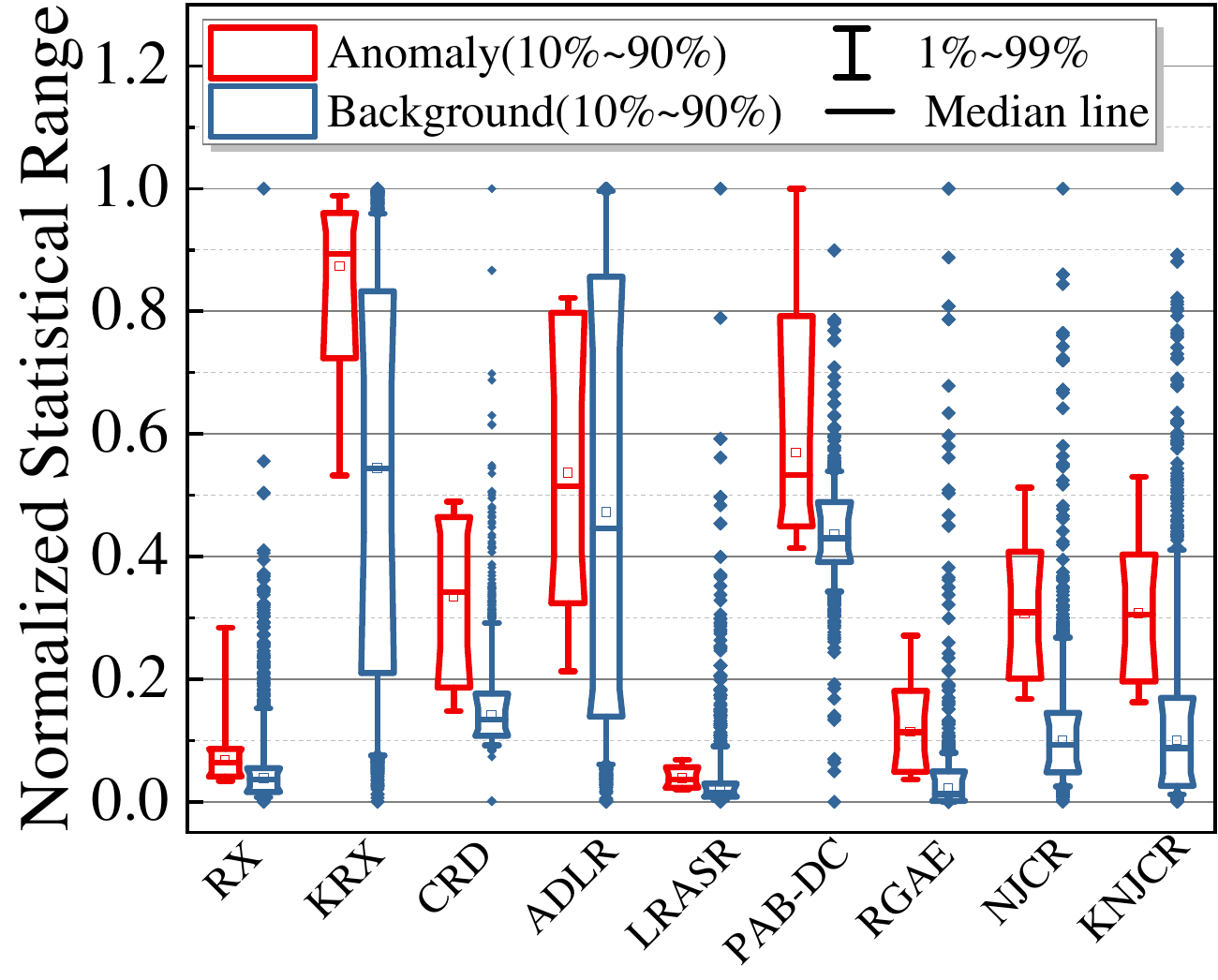}
}%

\subfigure[]{
\centering
\includegraphics[width=0.6\linewidth]{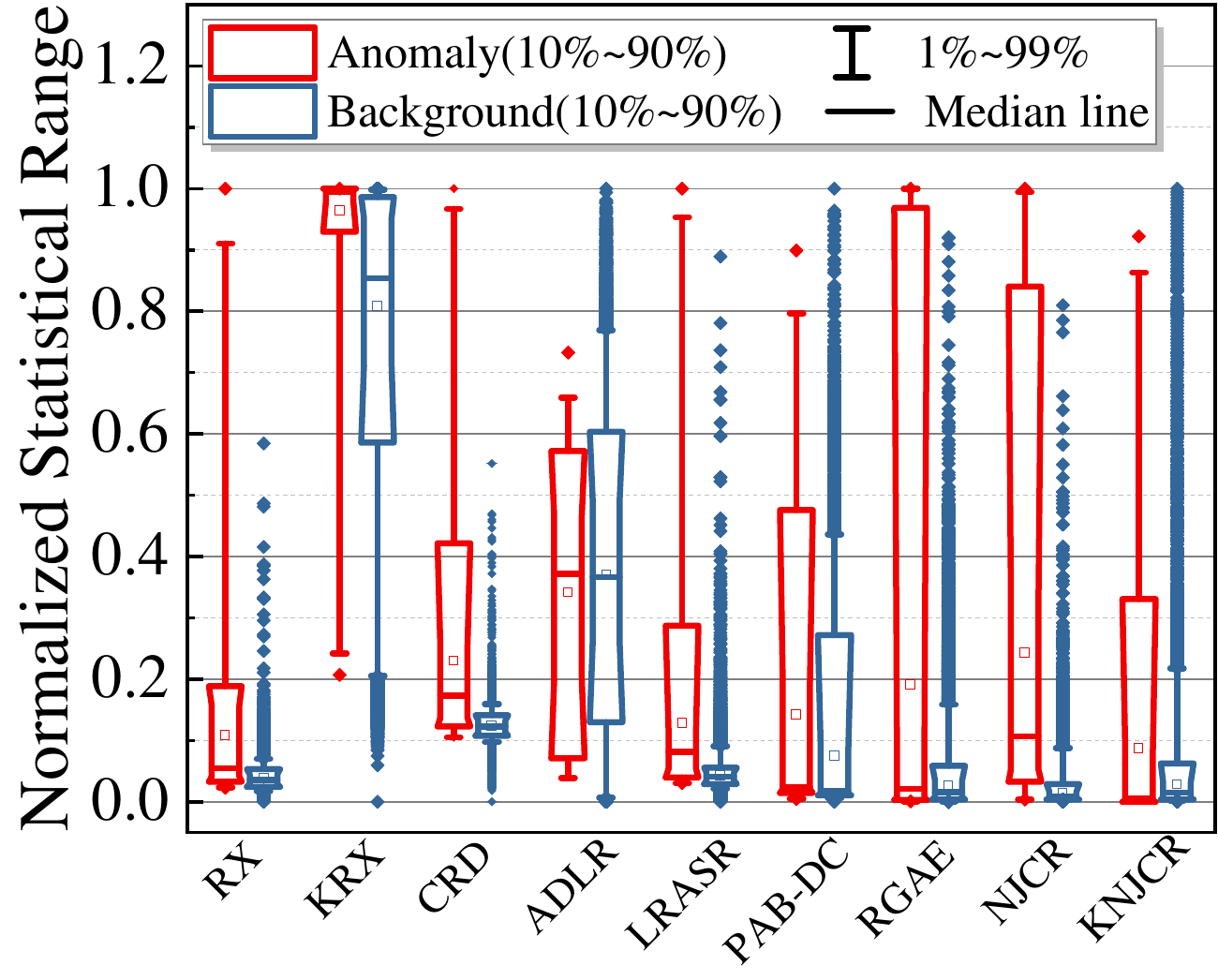}
}%

\subfigure[]{
\centering
\includegraphics[width=0.6\linewidth]{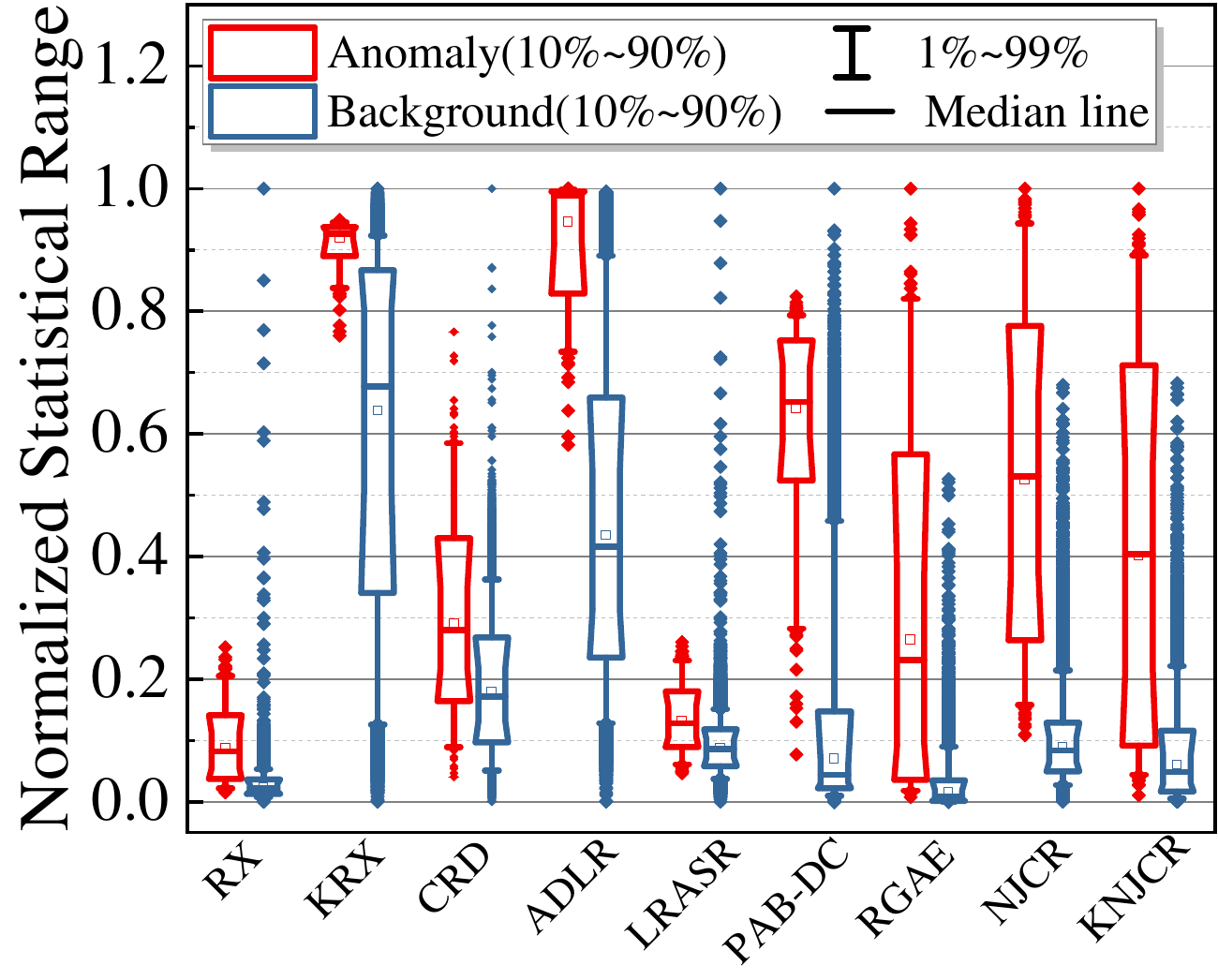}
}%
\centering
\caption{The background-target separability maps for four datasets. (a) LCVF. (b) SanDiego. (c) Viareggio. (d) CRi.}
\label{annotation 10}
\end{figure}

For the CRi dataset, Fig. \ref{annotation 9} shows that the visualized outputs of the anomalies and the backgrounds are not clearly detected by RX, CRD, and LRASR. Compared to KRX, ADLR, and PAB-DC, RGAE and the two proposed models obtain better visualization results between the two classes. Meanwhile, we can find that the visualized result of the background obtained by KNJCR is sharper than that of NJCR.

To better illustrate how anomalies are separated from backgrounds, we utilize the statistical range to evaluate the anomaly visibility enhancements. The pixels belonging to the background class and the anomaly class are extracted and show their statistical distributions in a box diagram. Since the detection results of different methods vary widely, we adopt a normalized detection result, which can be written as

\begin{equation*}
\tilde{R}(x)=\frac{R(x)-R_{min}(x)}{R_{max}(x)-R_{min}(x)},
\end{equation*}
where $R(x)$ is the detection result corresponding to $x$, $R_{min}(x)$ and $R_{max}(x)$ are the minimum and the maximum values of all pixels, respectively, and $\tilde{R}(x)$ is the normalized detection result.

The normalized background-anomaly statistical range of these methods for four datasets is shown in Fig. \ref{annotation 10}. The anomaly class is represented by the red box and the background class is represented by the blue one. The normalized values range between $10\%$ and $90\%$ is enclosed by the box, and the top line to the bottom line of each column represents the range of values between $1\%\sim99\%$. It is known that the positions of the background and anomaly boxes reflect the separability of the detector, so we expect to have a larger separability distance and a smaller background range.

\begin{figure}[tbp]
\centering
\subfigure[]{
\includegraphics[width=0.65\linewidth]{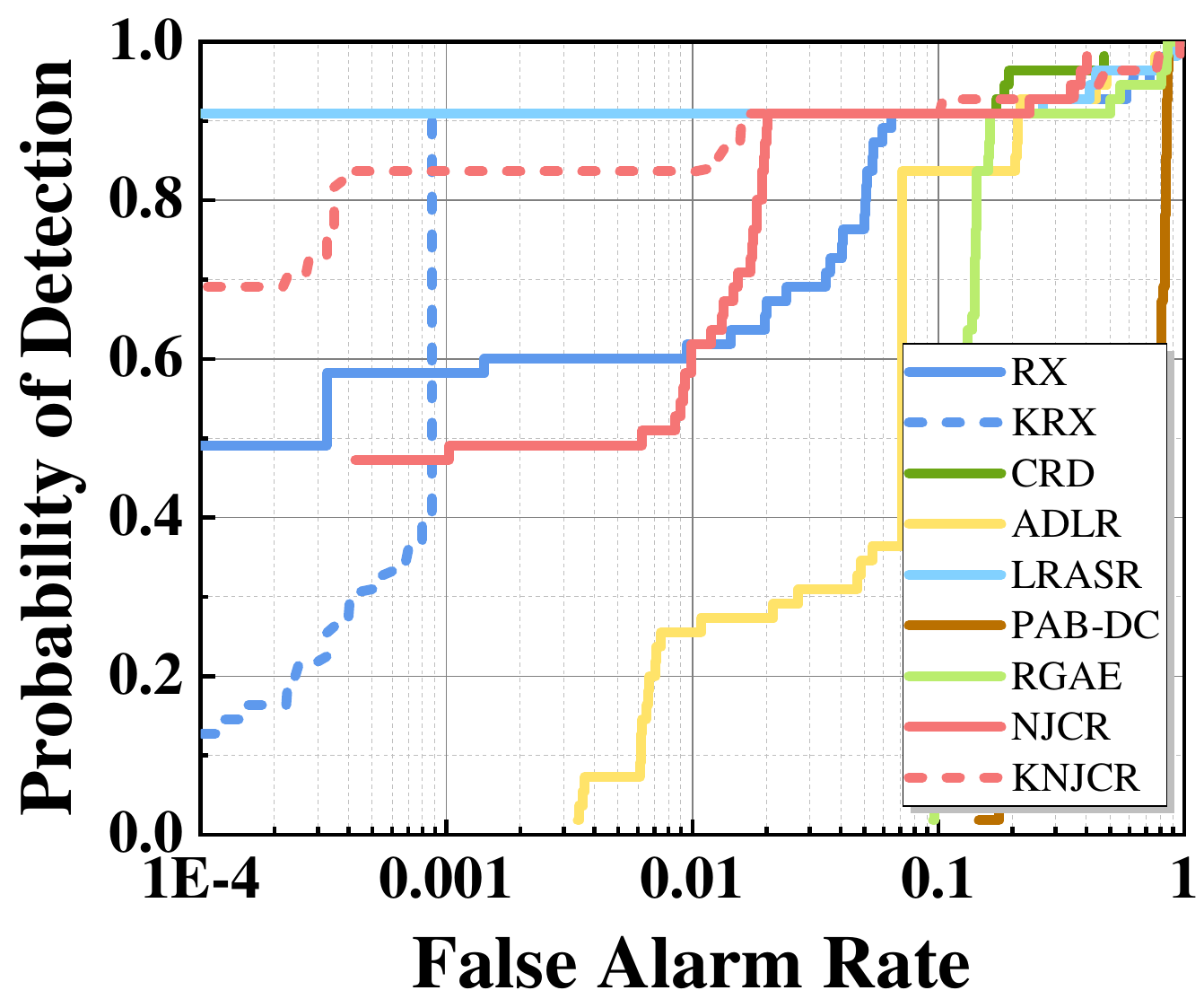}
}%

\subfigure[]{
\includegraphics[width=0.65\linewidth]{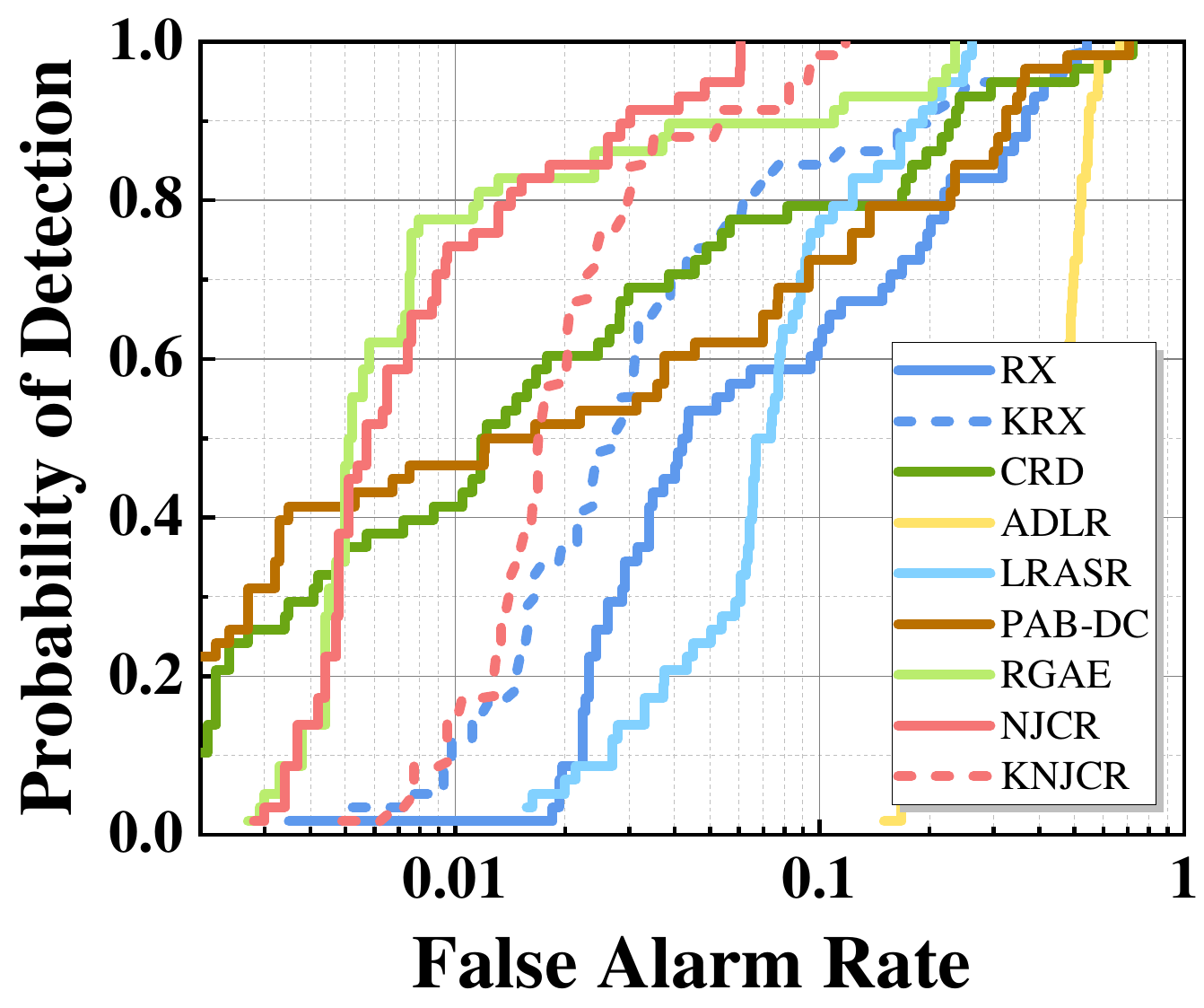}
}%

\subfigure[]{
\includegraphics[width=0.65\linewidth]{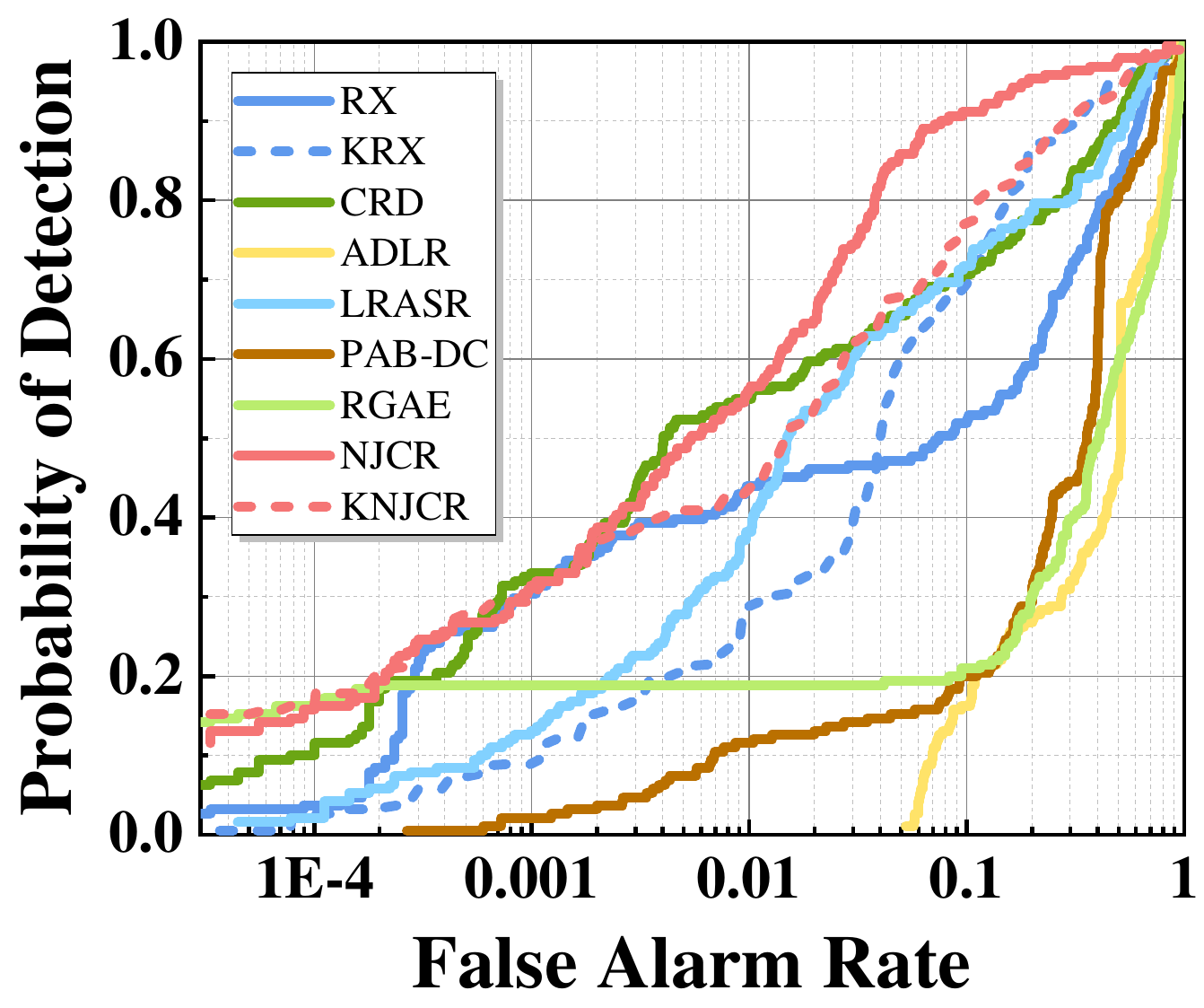}
}%

\subfigure[]{
\includegraphics[width=0.65\linewidth]{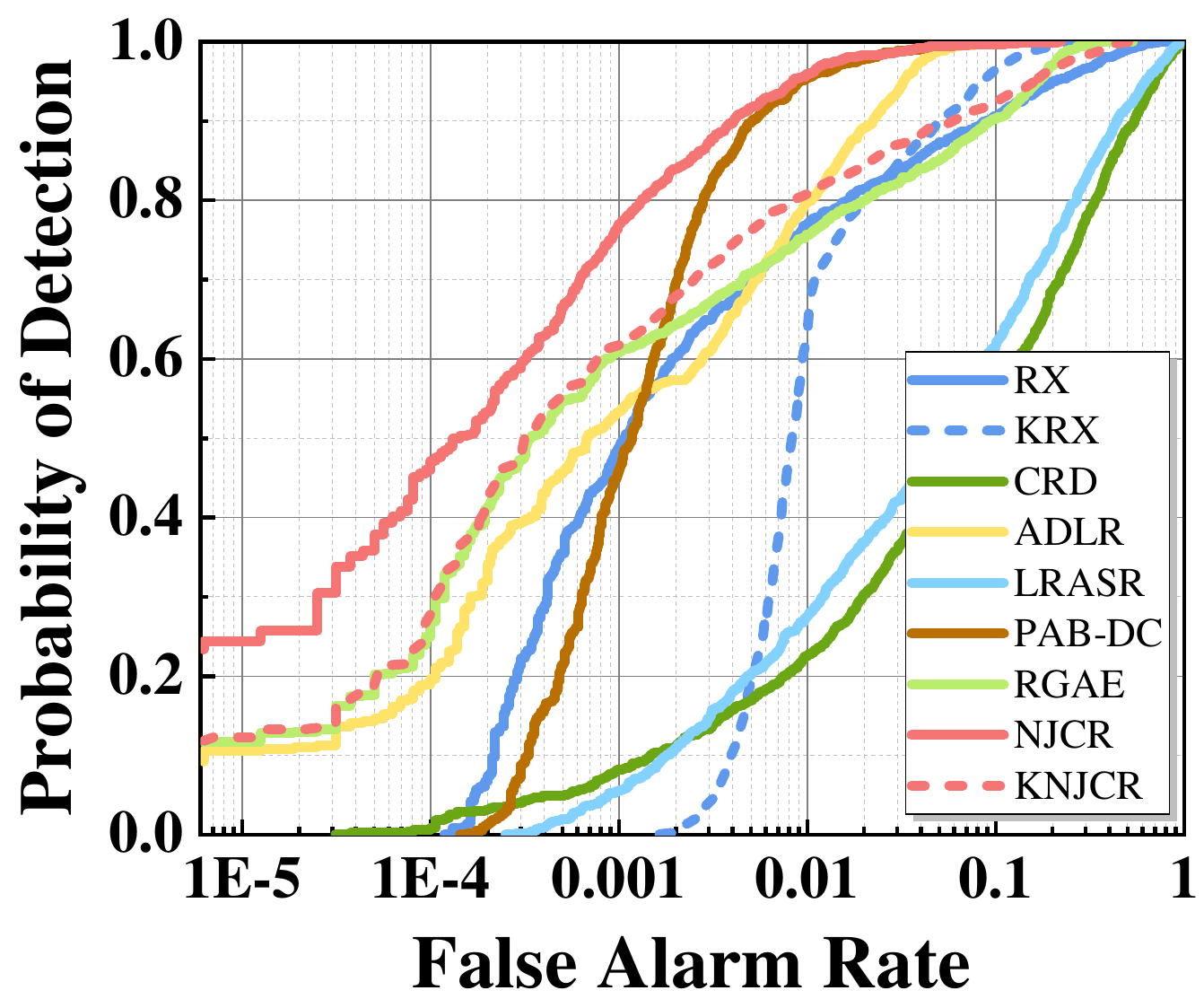}
}%
\centering
\caption{The ROC curves for four datasets. (a) LCVF. (b) SanDiego. (c) Viareggio. (d) CRi.}
\label{annotation 11}
\end{figure}

For the LCVF dataset, ADLR, PAB-DC, and RGAE failed to separate the enclosed part of two classes, and the background distribution range of KRX is relatively high. The background and anomaly boxes of RX, CRD, LRASR, NJCR, and KNJCR are separated, while the proposed NJCR model can better exclude anomalies with a smaller distribution range. For the SanDiego dataset, KRX and PAB-DC fail to separate the main part of the backgrounds and the anomalies. The gaps between the two boxes of RX, CRD, LRASR, and RGAE are very tight. The ADLR cannot correctly separate the anomaly class and the background class. Among all these methods, the proposed NJCR and KNJCR methods have relatively large separation gaps. Considering that the anomalies of the Viareggio dataset are more difficult to be detected, as shown in Fig. \ref{annotation 10} (c), only the proposed NJCR model can separate the middle $80\%$ values of the two classes. In contrast, the KNJCR models have smaller overlaps between the anomaly box and the background box. For the CRi dataset, KRX, ADLR, PAB-DC, and NJCR separate the backgrounds and the anomalies perfectly, while the proposed NJCR model can suppress the background information into a smaller distribution range.

\begin{table*}[tbp]
\caption{AUC values and time consumption of the comparable methods for four datasets.}
\centering
{\scriptsize
\begin{tabular}{c|ccc|ccc|ccc|ccc} 
\toprule
\multirow{2}{*}{Methods} &\multicolumn{3}{c|}{LCVF} &\multicolumn{3}{c|}{SanDiego} & \multicolumn{3}{c|}{Viareggio} &\multicolumn{3}{c}{CRi}\\  
 &$\text{AUC}_{(\text{Pf,Pd})}$&$\text{AUC}_{(\text{Pf,}\tau)}$&Time(s) &$\text{AUC}_{(\text{Pf,Pd})}$&$\text{AUC}_{(\text{Pf,}\tau)}$&Time(s) &$\text{AUC}_{(\text{Pf,Pd})}$&$\text{AUC}_{(\text{Pf,}\tau)}$&Time(s)
 &$\text{AUC}_{(\text{Pf,Pd})}$&$\text{AUC}_{(\text{Pf,}\tau)}$&Time(s)\\ \midrule
RX &0.9293&0.0630&1.05 &0.8520&0.1198&0.1 &0.7970&0.2006&6.67 &0.9677 &0.0319& 0.56\\
CRD &0.9144&\textbf{0.0272}&2200 &0.9096&0.0706&299.1 &0.8757&0.1218&3989 &0.8205&0.1791&4534\\
ADLR &0.8950&0.0969&49.71 &0.4906&0.4264&5.12 &0.5367&0.4606&75.12 &0.9933&0.0066&138.4\\
LRASR &0.8355&0.0444&1735 &0.8973&0.0844&597.5 &0.8658&0.1319&4159 &0.8542&0.1454&6462\\
PAB-DC &0.5156&0.4752&5677 &0.8878&0.0883&607.3 &0.6591&0.3383& 113723  &0.9970&0.0029& 44318 \\
RGAE &0.8205&0.1709&631.5 &0.9692&0.0239&163.9 &0.5560&0.4414&2800 &0.9756&0.0241&1359\\
NJCR &\textbf{0.9586}&0.0363&79.67 &\textbf{0.
9856} &\textbf{0.0115}&13.60  &\textbf{0.9541}&\textbf{0.0436}&216.9 &\textbf{0.9978}&\textbf{0.0021}&328.1\\
\midrule
KRX &0.9285&0.0511&1070 &0.9215&0.0598&79.85 &0.8877&0.1097&1018 &{\textbf{0.9804}}&{\textbf{0.0194}}&716.5\\
KNJCR &{\textbf{0.9621}}&{\textbf{0.0408}}&114.2  &{\textbf{0.9944}}&{\textbf{0.0306}}&41.52 &{\textbf{0.9151}}&{\textbf{0.0897}}&374.7 &0.9780&0.0313&419.1    \\
\bottomrule
\end{tabular}}
\label{Table2}
\end{table*}

For quantitative comparisons, the receiver operating characteristic (ROC) curves plotted by the probability of detection ($P_D$) and false alarm rate ($P_F$) are the most effective way to evaluate the detection efficacy of all algorithms. And the area under ROC curves (AUC) values have been used to further compare the performances. It is known that every pair of ($P_D$, $P_F$) is determined by a specific value of threshold $\tau$, thus, the AUC values calculated by ($P_F$, $\tau$) can also evaluate the effectiveness of the detectors on background suppression \cite{chang2020effective}. Effective methods are supposed to have relatively larger AUC values calculated by $(P_D, P_F)$ and smaller AUC values calculated by $(P_F, \tau)$ \cite{wang2021auto}.

In this article, the conventional ROC curves, the AUC$_{(P_D, P_F)}$, and the AUC$_{(P_F, \tau)}$ are utilized for quantitative analysis, as shown in Fig. \ref{annotation 11} and Table. \ref{Table2}, respectively. It can be found that in most cases, the proposed nonnegative-constrained joint collaborative representation models outperform other existing detectors. Among them, the proposed NJCR model yields the best ROC curves and the highest AUC$_{(P_D, P_F)}$ values on the SanDiego, Viareggio, and CRi datasets. And the AUC$_{(P_F, \tau)}$ values of the NJCR are obviously smaller than other detectors. In LCVF datasets, it can be seen that CRD and LRASR can quickly detect 90$\%$ of the anomalies when the false alarm rate is relatively low ($<1e-3$), while they cannot perfectly detect the remaining $10\%$ anomalies with low FARs. Since the inserted anomalies are small, some of them are subpixel-level, correctly detecting all pixels that contains abnormal information is highly important for this dataset. Combining the AUC values, we can recognize that the proposed NJCR methods have more balanced performances with higher AUCs.

For the kerneled detectors, the performance of KNJCR is better than KRX in the LCVF, SanDiego, and Viareggio datasets. In the CRi dataset, KRX outperforms the proposed KNJCR. The possible reasons are: (1) In some cases, the constructed global union dictionary is not suitable for the kerneled model; (2) the spectra differences between anomalies and backgrounds in these datasets are linearly separable and have hardly nonlinear correlations. We may thus conclude that directly mapping the CR model into a nonlinear feature space may not help improve the detection performances. Considering the computational complexity usually caused by computing kernels, the original NJCR model is more appropriate for the proposed anomaly detection task.

Furthermore, the processing time of all algorithms is also recorded and shown in Table. \ref{Table2}. We can see that the consuming time of the proposed NJCR and KNJCR models are much smaller than other representation-based models, such as CRD, LRASR, and PAB-DC. The good performance of PAB-DC in the CRi dataset is compromised with huge computational requirements and time consumption. On the other hand, the detection performances of RX and ADLR, whose processing time is quite fast, are not satisfactory. By Taking an overall consideration of the computation time and detection performances, the proposed NJCR and KNJCR models are more competitive for the anomaly detection task.
\subsection{Parametric Analysis}
This section primarily discusses how detection performances are affected by the parameters of the proposed NJCR and KNJCR objectives in the four HSI datasets used for testing and experimentation.

\begin{figure}[tbp]
\centering
\includegraphics[width=2.3in]{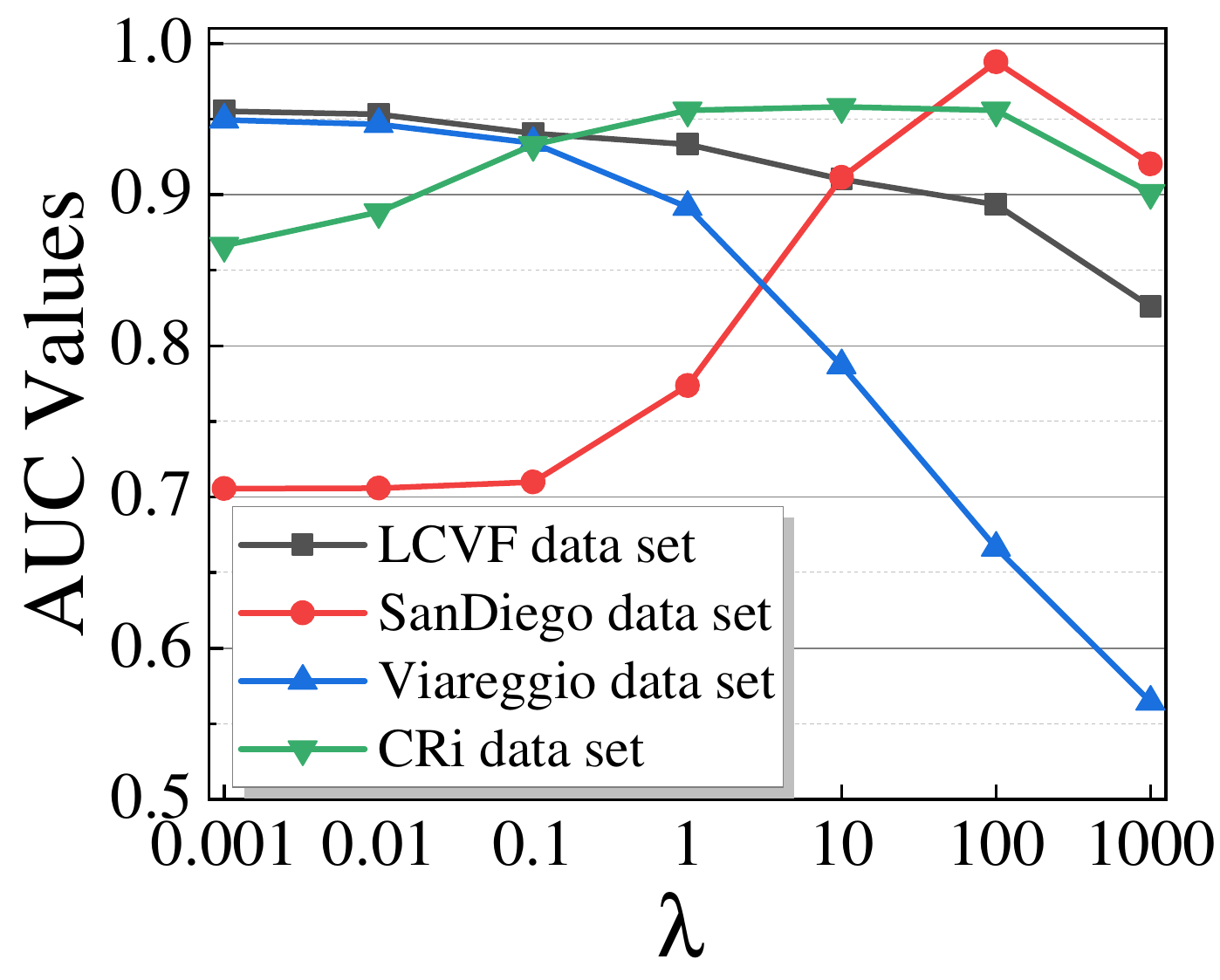}
\caption{The parametric analysis of NJCR for four datasets. (a) LCVF. (b) SanDiego. (c) Viareggio. (d) CRi.}
\label{annotation 12}
\end{figure}

The parameter of the NJCR model that is evaluated is the regularized scalar $\lambda$. Fig. \ref{annotation 12} shows the AUC values of NJCR model when $\lambda$ is set from $0.001$ to $1000$. It can be seen that as $\lambda$ increases, the detection efficacy of the NJCR model in LCVF  and  Viareggio dataset decreases,  and that $\lambda=0.001$ is optimum. When the value of $\lambda$ is less than 1, the detection efficacy is not much influenced.  For both the  SanDiego and  CRi datasets,  the  AUC values of the NJCR model increase at first and then slightly decrease when $\lambda$ is greater than 100. The suggested value of $\lambda$ is $100$. We may derive an interesting conclusion that the suggested values of $\lambda$ for the NJCR model are influenced by the sizes of anomalies. For subpixel-level detection tasks, like what we encounter in the LCVF and Viareggio datasets, $\lambda$ needs to be small. And for larger target detection tasks, the value of $\lambda$ is larger.

\begin{figure}[tbp]
\centering
\subfigure[]{
\includegraphics[width=0.72\linewidth]{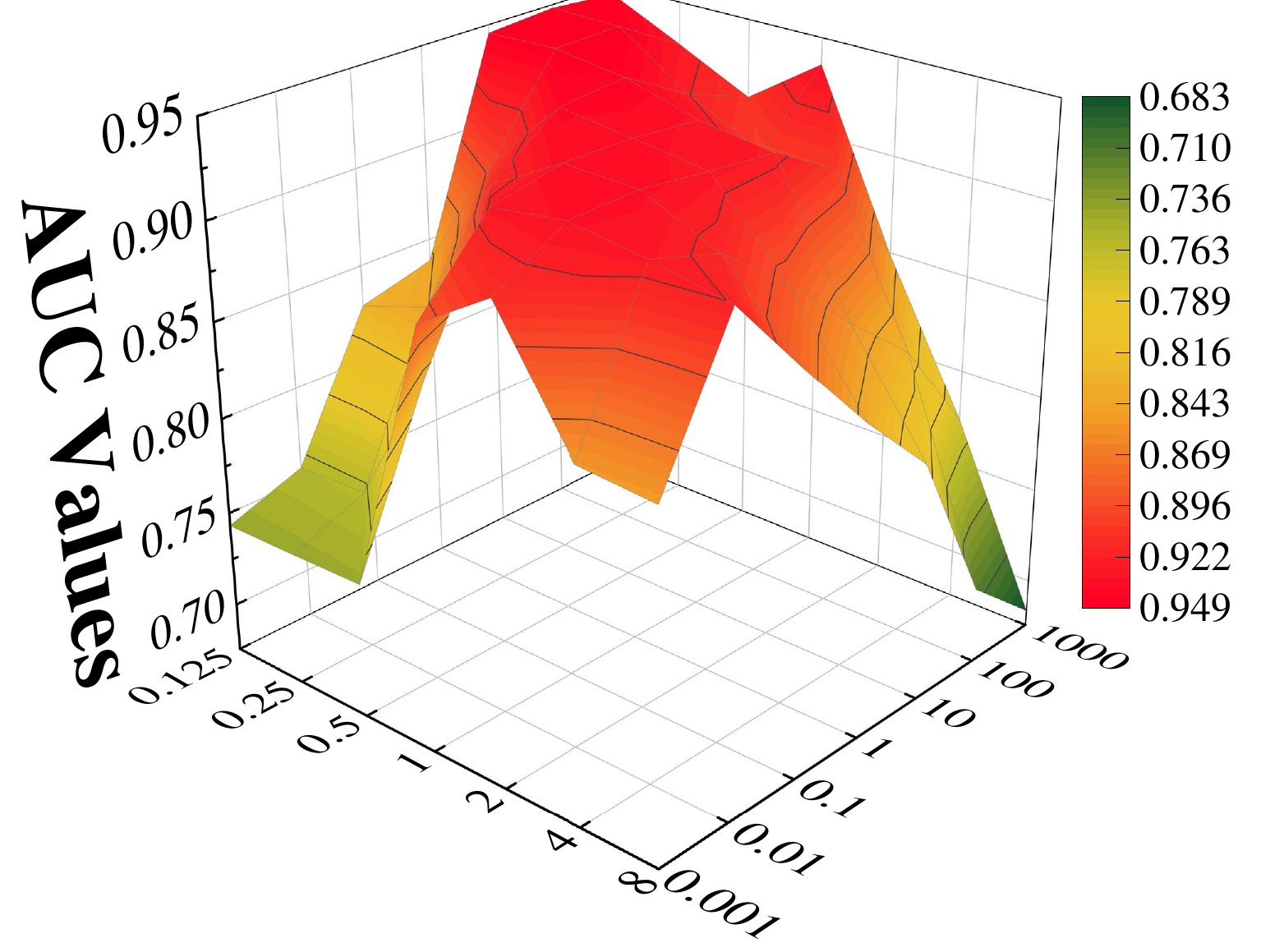}
}%

\subfigure[]{
\includegraphics[width=0.72\linewidth]{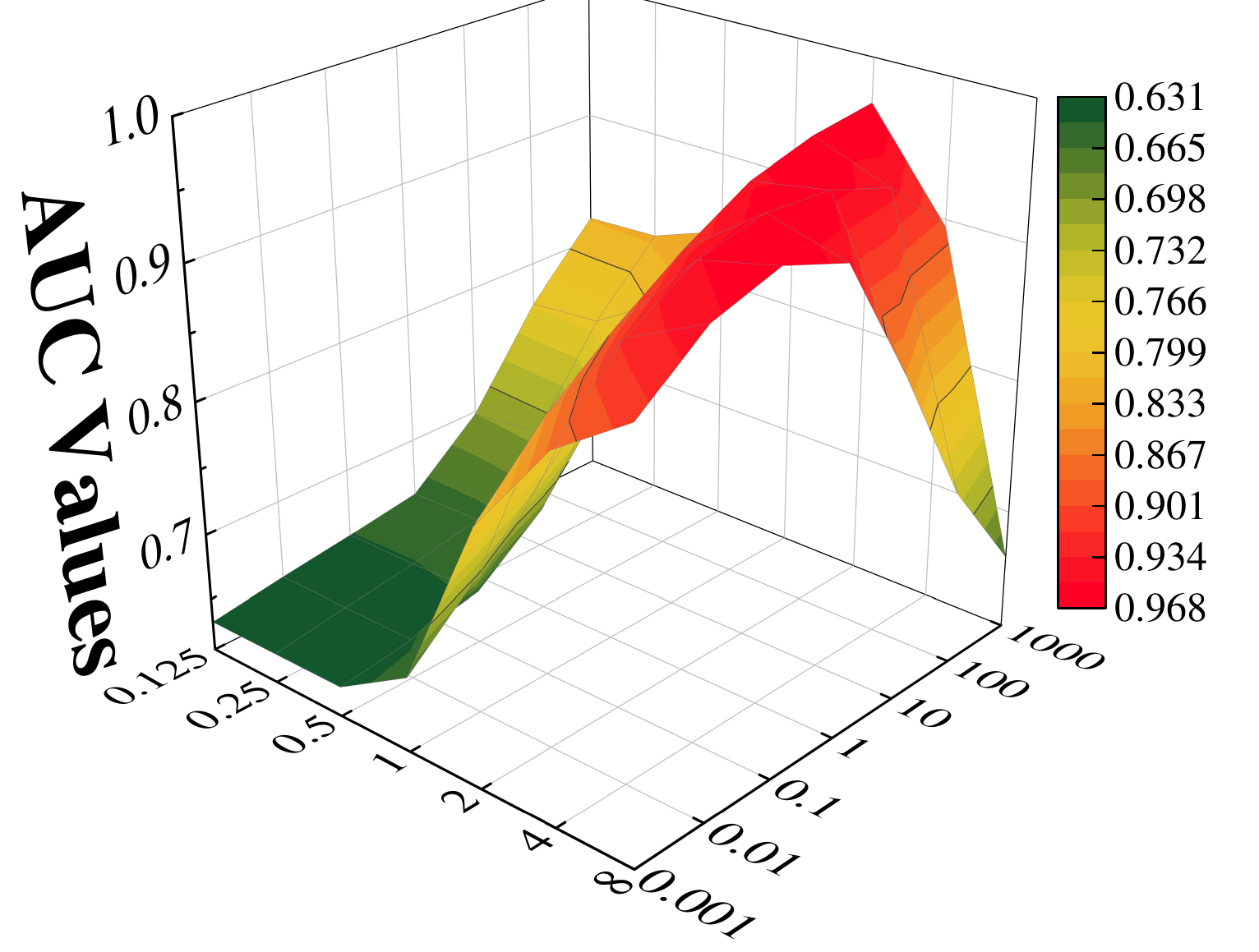}
}%

\subfigure[]{
\includegraphics[width=0.72\linewidth]{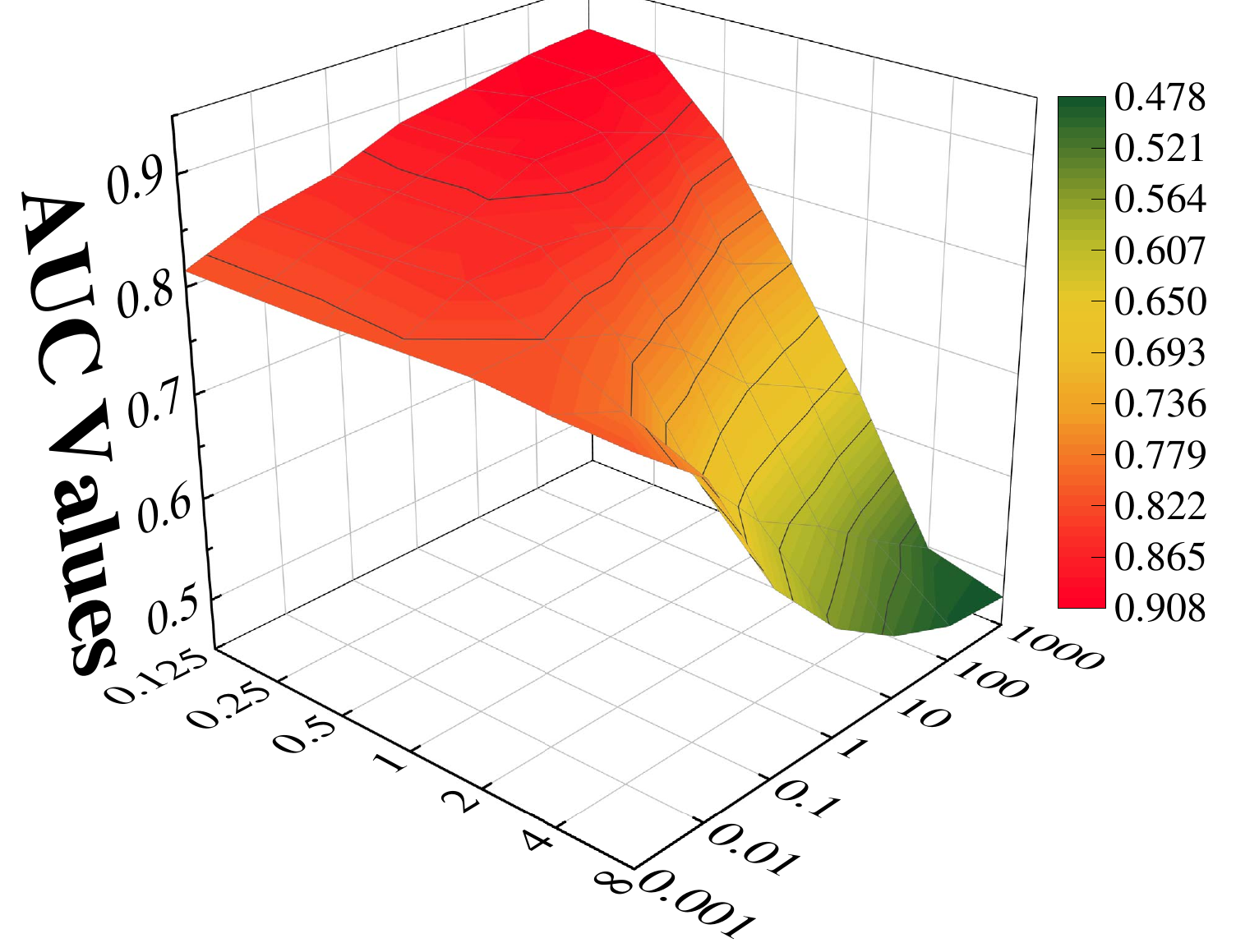}
}%

\subfigure[]{
\includegraphics[width=0.72\linewidth]{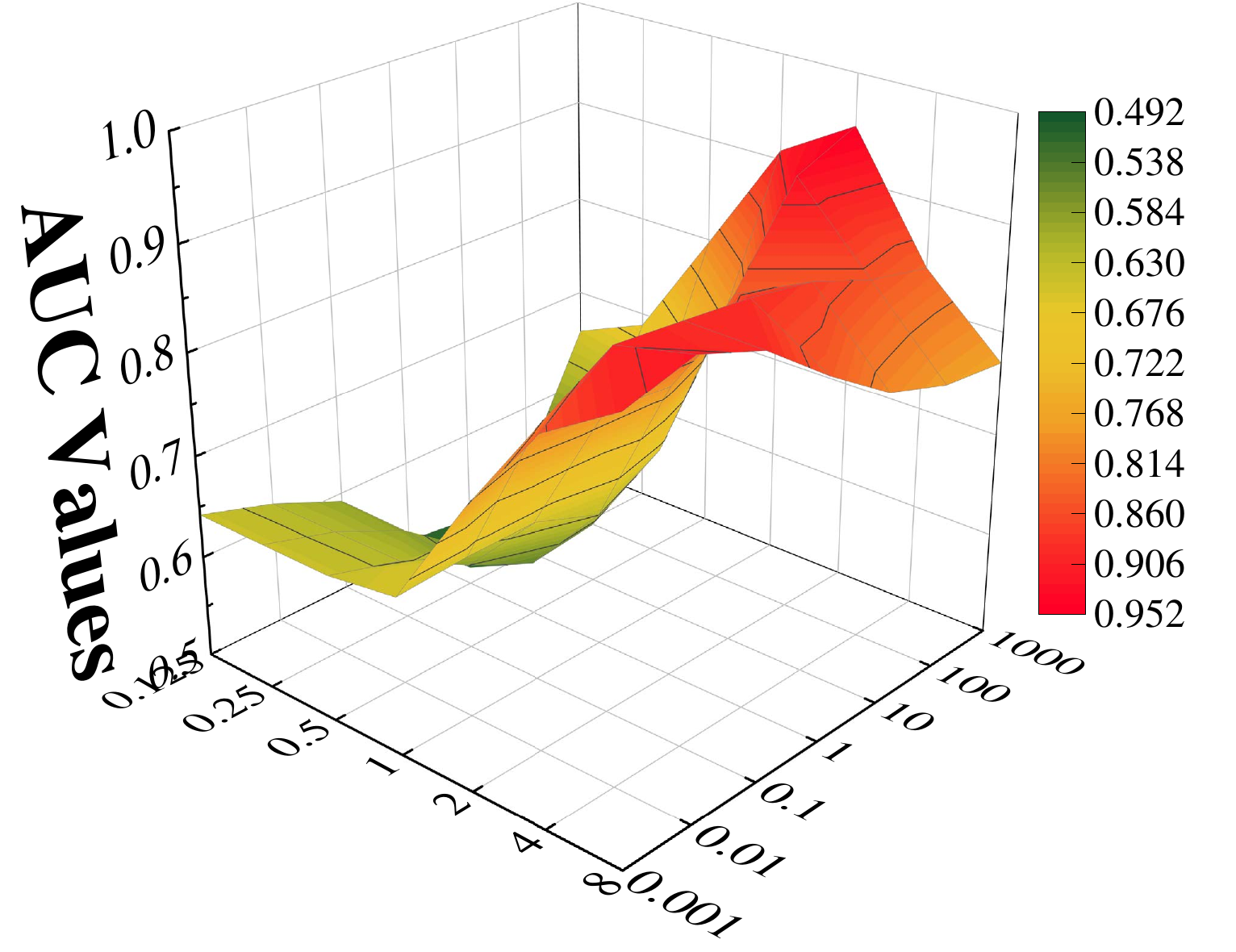}
}%

\centering
\caption{The parametric analysis of KNJCR for four datasets. (a) LCVF. (b) SanDiego. (c) Viareggio. (d) CRi.}
\label{annotation 13}
\end{figure}

For the KNJCR model, we made a joint evaluation of the kernel width $\sigma$ and the regularized scalar $\lambda$.  The AUC values of KNJCR under different parametric settings are shown in Fig. \ref{annotation 13} with a 3D colored surface map. It can be seen that the optimal values of $\lambda$ are partly influenced by the value of $\sigma$. Interestingly, these two parameters are similarly negatively correlated in all four datasets: the combination of a larger $\sigma$ and smaller $\lambda$, or a smaller $\sigma$ and larger $\lambda$ each has better AUCs. Experience shows that the value of kernel width $\sigma$ can be set in the range of $[0.25, 0.5]$ for the LCVF and Viareggio datasets and 4 for the SanDiego and CRi datasets. And the regularized scalar $\lambda$ can be set to 100.
\subsection{Ablation Study}
In this section, a detailed ablation study is conducted on the LCVF dataset to further investigate the efficacy of the proposed models. 

First of all, to test the effect of the constraints added to the objective function, we evaluate the performances of the proposed models concerning the sum-to-one constraint and the nonnegative constraint. As shown in Table \ref{Table3}, the AUC values are calculated by the joint CR model, the nonnegative constrained objective, the sum-to-one constrained objective, and with both nonnegative and sum-to-one constrained objectives, respectively. It can be seen that both constraints are helpful in terms of improving the performances of the model compare to the original joint CR model, and with a fully constrained version, both NJCR and KNJCR, the model reaches the best detection performance. 

Furthermore, we also give an analysis about the influences of the union dictionary and the Frobenius norm for the joint collaborative representation models. The AUC values of two models under these settings are also illustrated in Table \ref{Table3}. It can be clearly seen that the union dictionary can obviously improve the detection accuracy of the proposed models. By means of the unified background and anomaly dictionary, the spectral information belonging to the background class and the anomaly class is separated and will return a more separable result after calculating the residual of the CR model. Thus, the AUC values are obviously improved. It is known that the original CR model optimizes the reconstruction error pixel-by-pixel, thus, the objectives of NJCR and KNJCR are solved by the $\ell_2$-minimization process. The time consumption of the original CR model corresponding to NJCR and KNJCR are 774.6s and 962.1s, respectively. We can find that both the time consumption and detection performance are greatly improved by using the designed Frobenius norm.

\begin{table}[tbp]
\caption{Ablation Study of the Proposed Models.}
\centering %
\begin{tabular}{c|cc} 
\toprule
   & NCR    & KNCR   \\ \midrule
Joint CR Objective     & 0.7543 & 0.8982 \\
Nonnegative Constrained & 0.7992 & 0.8979 \\
Sum-to-one Constrained  & 0.9194 & 0.8929 \\
Background sub-dictionary  & 0.6614 & 0.8633 \\
$\ell_2$-norm                & 0.7867 & 0.9304 \\
All Constrained        & \textbf{0.9586} & \textbf{0.9504} \\
\bottomrule
\end{tabular}
\label{Table3}
\end{table}
\section{Conclusion}
In this paper, a novel joint collaborative representation model is explored by designing nonnegative constraints and constructing a global union dictionary. Unlike previous representation-based detectors, the proposed NJCR models assume that the coefficients are nonnegative, obey the sum-to-one rule, and adopt the Frobenius norm to jointly optimize the whole coefficient matrix of the image. To better approximate the signals, a unified background and anomaly dictionary is constructed, and the final result is obtained by calculating residuals that exclude the background information. Furthermore, a kernel version is proposed for nonlinear analysis. To verify the effectiveness of the proposed methods, extensive experiments have been conducted comparing the proposed model with seven state-of-the-art detectors. The original NJCR model shows superior results by providing effective separations between the backgrounds and the anomalies, and the kerneled NJCR model also shows great performances compared to the nonlinear detector KRX and other comparative algorithms. Compare to classic representation-based methods, our method can achieve good performance while saving time consumption. The ablation study further emphasizes that the Frobenius norm, sum-to-one constraint, and nonnegative constraint have a great impact on improving the detection performance. Through the experiments, we have noticed that the preferred value of $\lambda$ varies in different datasets, which may be influenced by the size of anomalies. We will keep finding possible proofs by conducting more experimental research and mathematical analysis in the further. We have also noticed that directly mapping the collaborative representation model into a high dimensional feature space may not necessarily improve the efficacy of signal recovery, and thus, result in worse performances. Therefore, we will persist in our study to construct precise dictionaries and design more accurate models to better approximate the HSI data and separate the binary classes.

\section*{Acknowledgment}
The authors would like to thank the handling editor and
the anonymous reviewers for their careful reading and helpful
remarks, the authors of NCut, CRD, ADLR, LRASR, PAB-DC, and RGAE algorithms for sharing their codes.

\ifCLASSOPTIONcaptionsoff
\newpage
\fi
\bibliographystyle{IEEEtran}
\bibliography{ref.bib}

\begin{thebibliography}{10}
\providecommand{\url}[1]{#1}
\csname url@samestyle\endcsname
\providecommand{\newblock}{\relax}
\providecommand{\bibinfo}[2]{#2}
\providecommand{\BIBentrySTDinterwordspacing}{\spaceskip=0pt\relax}
\providecommand{\BIBentryALTinterwordstretchfactor}{4}
\providecommand{\BIBentryALTinterwordspacing}{\spaceskip=\fontdimen2\font plus
\BIBentryALTinterwordstretchfactor\fontdimen3\font minus
  \fontdimen4\font\relax}
\providecommand{\BIBforeignlanguage}[2]{{%
\expandafter\ifx\csname l@#1\endcsname\relax
\typeout{** WARNING: IEEEtran.bst: No hyphenation pattern has been}%
\typeout{** loaded for the language `#1'. Using the pattern for}%
\typeout{** the default language instead.}%
\else
\language=\csname l@#1\endcsname
\fi
#2}}
\providecommand{\BIBdecl}{\relax}
\BIBdecl

\bibitem{7882742}
P.~Ghamisi, J.~Plaza, Y.~Chen, J.~Li, and A.~J. Plaza, ``Advanced spectral
  classifiers for hyperspectral images: A review,'' \emph{IEEE Trans. Geosci.
  Remote Sens.}, vol.~5, no.~1, pp. 8--32, 2017.

\bibitem{9186822}
S.~Liu, Q.~Shi, and L.~Zhang, ``Few-shot hyperspectral image classification
  with unknown classes using multitask deep learning,'' \emph{IEEE Trans. Geos.
  Remote Sens.}, vol.~59, no.~6, pp. 5085--5102, 2021.

\bibitem{WAMBUGU2021102603}
N.~Wambugu, Y.~Chen, Z.~Xiao, K.~Tan, M.~Wei, X.~Liu, and J.~Li,
  ``Hyperspectral image classification on insufficient-sample and feature
  learning using deep neural networks: A review,'' \emph{Int. J. Appl. Earth
  Obs. Geoinf.}, vol. 105, p. 102603, 2021.

\bibitem{matteoli2010tutorial}
S.~Matteoli, M.~Diani, and G.~Corsini, ``A tutorial overview of anomaly
  detection in hyperspectral images,'' \emph{IEEE Aerosp. Electron. Syst.
  Mag.}, vol.~25, no.~7, pp. 5--28, 2010.

\bibitem{9532003}
H.~Su, Z.~Wu, H.~Zhang, and Q.~Du, ``Hyperspectral anomaly detection: A
  survey,'' \emph{IEEE Trans. Geosci. Remote Sens.}, pp. 2--28, 2021.

\bibitem{zou2017random}
Z.~Zou and Z.~Shi, ``Random access memories: A new paradigm for target
  detection in high resolution aerial remote sens. images,'' \emph{IEEE Trans.
  Image Process.}, vol.~27, no.~3, pp. 1100--1111, 2017.

\bibitem{fowler2011anomaly}
J.~E. Fowler and Q.~Du, ``Anomaly detection and reconstruction from random
  projections,'' \emph{IEEE Trans. Image Process.}, vol.~21, no.~1, pp.
  184--195, 2011.

\bibitem{nasrabadi2013hyperspectral}
N.~M. Nasrabadi, ``Hyperspectral target detection: An overview of current and
  future challenges,'' \emph{IEEE Signal Process. Mag.}, vol.~31, no.~1, pp.
  34--44, 2013.

\bibitem{stein2002anomaly}
D.~W. Stein, S.~G. Beaven, L.~E. Hoff, E.~M. Winter, A.~P. Schaum, and A.~D.
  Stocker, ``Anomaly detection from hyperspectral imagery,'' \emph{IEEE Signal
  Process. Mag.}, vol.~19, no.~1, pp. 58--69, 2002.

\bibitem{reed1990adaptive}
I.~S. Reed and X.~Yu, ``Adaptive multiple-band {CFAR} detection of an optical
  pattern with unknown spectral distribution,'' \emph{IEEE Transactions on
  Acoustics, Speech, and Signal Processing}, vol.~38, no.~10, pp. 1760--1770,
  1990.

\bibitem{borghys2011hyperspectral}
D.~Borghys, V.~Achard, S.~Rotman, N.~Gorelik, C.~Perneel, and E.~Schweicher,
  ``Hyperspectral anomaly detection: A comparative evaluation of methods,'' in
  \emph{2011 XXXth URSI General Assembly and Scientific Symposium}.\hskip 1em
  plus 0.5em minus 0.4em\relax IEEE, 2011, pp. 1--4.

\bibitem{billor2000bacon}
N.~Billor, A.~S. Hadi, and P.~F. Velleman, ``{BACON}: blocked adaptive
  computationally efficient outlier nominators,'' \emph{Computational
  statistics \& data analysis}, vol.~34, no.~3, pp. 279--298, 2000.

\bibitem{du2010random}
B.~Du and L.~Zhang, ``Random-selection-based anomaly detector for hyperspectral
  imagery,'' \emph{IEEE Trans. Geos. Remote Sens.}, vol.~49, no.~5, pp.
  1578--1589, 2010.

\bibitem{amiel2020consensus}
Y.~Amiel, A.~Frajman, and S.~R. Rotman, ``Consensus anomaly detection using
  clustering methods in hyperspectral imagery,'' in \emph{Imaging Spectrometry
  XXIV: Applications, Sensors, and Processing}, vol. 11504.\hskip 1em plus
  0.5em minus 0.4em\relax SPIE, 2020, pp. 71--84.

\bibitem{adler2009improved}
S.~M. Adler-Golden, ``Improved hyperspectral anomaly detection in heavy-tailed
  backgrounds,'' in \emph{2009 First Workshop on Hyperspectral Image and Signal
  Processing: Evolution in Remote Sens.}\hskip 1em plus 0.5em minus 0.4em\relax
  IEEE, 2009, pp. 1--4.

\bibitem{veracini2010spectral}
T.~Veracini, S.~Matteoli, M.~Diani, G.~Corsini, and S.~U. de~Ceglie, ``A
  spectral anomaly detector in hyperspectral images based on a non-gaussian
  mixture model,'' in \emph{2010 2nd Workshop on Hyperspectral Image and Signal
  Processing: Evolution in Remote Sens.}\hskip 1em plus 0.5em minus 0.4em\relax
  IEEE, 2010, pp. 1--4.

\bibitem{schweizer2000hyperspectral}
S.~M. Schweizer and J.~M. Moura, ``Hyperspectral imagery: Clutter adaptation in
  anomaly detection,'' \emph{IEEE Trans. Inf. Theory}, vol.~46, no.~5, pp.
  1855--1871, 2000.

\bibitem{kwon2005kernel}
H.~Kwon and N.~M. Nasrabadi, ``Kernel {RX}-algorithm: A nonlinear anomaly
  detector for hyperspectral imagery,'' \emph{IEEE Trans. Geos. Remote Sens.},
  vol.~43, no.~2, pp. 388--397, 2005.

\bibitem{banerjee2006support}
A.~Banerjee, P.~Burlina, and C.~Diehl, ``A support vector method for anomaly
  detection in hyperspectral imagery,'' \emph{IEEE Trans. Geos. Remote Sens.},
  vol.~44, no.~8, pp. 2282--2291, 2006.

\bibitem{zhao2014robust}
R.~Zhao, B.~Du, and L.~Zhang, ``A robust nonlinear hyperspectral anomaly
  detection approach,'' \emph{IEEE J. Sel. Top. Appl. Earth Obs. Remote Sens.},
  vol.~7, no.~4, pp. 1227--1234, 2014.

\bibitem{scholkopf1997kernel}
B.~Sch{\"o}lkopf, A.~Smola, and K.-R. M{\"u}ller, ``Kernel principal component
  analysis,'' in \emph{International conference on artificial neural
  networks}.\hskip 1em plus 0.5em minus 0.4em\relax Springer, 1997, pp.
  583--588.

\bibitem{matteoli2013background}
S.~Matteoli, T.~Veracini, M.~Diani, and G.~Corsini, ``Background density
  nonparametric estimation with data-adaptive bandwidths for the detection of
  anomalies in multi-hyperspectral imagery,'' \emph{IEEE Geoscience and remote
  sensing letters}, vol.~11, no.~1, pp. 163--167, 2013.

\bibitem{arisoy2021nonparametric}
S.~Arisoy and K.~Kayabol, ``Nonparametric bayesian background estimation for
  hyperspectral anomaly detection,'' \emph{Digital Signal Processing}, vol.
  111, p. 102993, 2021.

\bibitem{zhao2017hyperspectral}
R.~Zhao, B.~Du, and L.~Zhang, ``Hyperspectral anomaly detection via a sparsity
  score estimation framework,'' \emph{IEEE Trans. Geos. Remote Sens.}, vol.~55,
  no.~6, pp. 3208--3222, 2017.

\bibitem{li2014collaborative}
W.~Li and Q.~Du, ``Collaborative representation for hyperspectral anomaly
  detection,'' \emph{IEEE Trans. Geos. Remote Sens.}, vol.~53, no.~3, pp.
  1463--1474, 2014.

\bibitem{rs11111318}
K.~Tan, Z.~Hou, F.~Wu, Q.~Du, and Y.~Chen, ``Anomaly detection for
  hyperspectral imagery based on the regularized subspace method and
  collaborative representation,'' \emph{Remote Sens.}, vol.~11, no.~11, 2019.

\bibitem{xu2015anomaly}
Y.~Xu, Z.~Wu, J.~Li, A.~Plaza, and Z.~Wei, ``Anomaly detection in hyperspectral
  images based on low-rank and sparse representation,'' \emph{IEEE Trans. Geos.
  Remote Sens.}, vol.~54, no.~4, pp. 1990--2000, 2015.

\bibitem{zhang2015low}
Y.~Zhang, B.~Du, L.~Zhang, and S.~Wang, ``A low-rank and sparse matrix
  decomposition-based mahalanobis distance method for hyperspectral anomaly
  detection,'' \emph{IEEE Trans. Geos. Remote Sens.}, vol.~54, no.~3, pp.
  1376--1389, 2015.

\bibitem{li2020low}
L.~Li, W.~Li, Q.~Du, and R.~Tao, ``Low-rank and sparse decomposition with
  mixture of gaussian for hyperspectral anomaly detection,'' \emph{IEEE Trans.
  Cybern.}, vol.~51, no.~9, pp. 4363--4372, 2020.

\bibitem{cheng2019graph}
T.~Cheng and B.~Wang, ``Graph and total variation regularized low-rank
  representation for hyperspectral anomaly detection,'' \emph{IEEE Trans. Geos.
  Remote Sens.}, vol.~58, no.~1, pp. 391--406, 2019.

\bibitem{su2020low}
H.~Su, Z.~Wu, A.-X. Zhu, and Q.~Du, ``Low rank and collaborative representation
  for hyperspectral anomaly detection via robust dictionary construction,''
  \emph{ISPRS J. Photogramm. Remote Sens.}, vol. 169, pp. 195--211, 2020.

\bibitem{Su2018Hyperspectral}
H.~Su, Z.~Wu, Q.~Du, and P.~Du, ``Hyperspectral anomaly detection using
  collaborative representation with outlier removal,'' \emph{IEEE J. Sel. Top.
  Appl. Earth Obs. Remote Sens.}, vol.~11, no.~12, pp. 5029--5038, 2018.

\bibitem{ZGY2020Hyperpsectral}
G.~Zhang, N.~Li, B.~Tu, Z.~Liao, and Y.~Peng, ``Hyperspectral anomaly detection
  via dual collaborative representation,'' \emph{IEEE J. Sel. Top. Appl. Earth
  Obs. Remote Sens.}, vol.~13, pp. 4881--4894, 2020.

\bibitem{6693730}
J.~Li, H.~Zhang, L.~Zhang, X.~Huang, and L.~Zhang, ``Joint collaborative
  representation with multitask learning for hyperspectral image
  classification,'' \emph{IEEE Trans. Geos. Remote Sens.}, vol.~52, no.~9, pp.
  5923--5936, 2014.

\bibitem{qu2018hyperspectral}
Y.~Qu, W.~Wang, R.~Guo, B.~Ayhan, C.~Kwan, S.~Vance, and H.~Qi, ``Hyperspectral
  anomaly detection through spectral unmixing and dictionary-based low-rank
  decomposition,'' \emph{IEEE Trans. Geos. Remote Sens.}, vol.~56, no.~8, pp.
  4391--4405, 2018.

\bibitem{9133150}
T.~Cheng and B.~Wang, ``Total variation and sparsity regularized decomposition
  model with union dictionary for hyperspectral anomaly detection,'' \emph{IEEE
  Trans. Geos. Remote Sens.}, vol.~59, no.~2, pp. 1472--1486, 2021.

\bibitem{6200362}
J.~M. Bioucas-Dias, A.~Plaza, N.~Dobigeon, M.~Parente, Q.~Du, P.~Gader, and
  J.~Chanussot, ``Hyperspectral unmixing overview: Geometrical, statistical,
  and sparse regression-based approaches,'' \emph{IEEE J. Sel. Top. Appl. Earth
  Obs. Remote Sens.}, vol.~5, no.~2, pp. 354--379, 2012.

\bibitem{wang2013kernel}
B.~Wang, W.~Li, N.~Poh, and Q.~Liao, ``Kernel collaborative
  representation-based classifier for face recognition,'' in \emph{Proc. -
  ICASSP IEEE Int. Conf. Acoust. Speech Signal Process.}, 2013, pp. 2877--2881.

\bibitem{shi2011face}
Q.~Shi, A.~Eriksson, A.~Van Den~Hengel, and C.~Shen, ``Is face recognition
  really a compressive sensing problem?'' in \emph{Proc. IEEE Comput. Soc.
  Conf. Comput. Vis. Pattern Recognit.}, 2011, pp. 553--560.

\bibitem{7097693}
J.~Li, H.~Zhang, and L.~Zhang, ``Efficient superpixel-level multitask joint
  sparse representation for hyperspectral image classification,'' \emph{IEEE
  Trans. Geos. Remote Sens.}, vol.~53, no.~10, pp. 5338--5351, 2015.

\bibitem{shi2000normalized}
J.~Shi and J.~Malik, ``Normalized cuts and image segmentation,'' \emph{IEEE
  Trans. Pattern Anal. Mach. Intell.}, vol.~22, no.~8, pp. 888--905, 2000.

\bibitem{felzenszwalb2004efficient}
P.~F. Felzenszwalb and D.~P. Huttenlocher, ``Efficient graph-based image
  segmentation,'' \emph{Int. J. Comput. Vis.}, vol.~59, no.~2, pp. 167--181,
  2004.

\bibitem{liu2011entropy}
M.-Y. Liu, O.~Tuzel, S.~Ramalingam, and R.~Chellappa, ``Entropy rate superpixel
  segmentation,'' in \emph{Proc. IEEE Comput. Soc. Conf. Comput. Vis. Pattern
  Recognit.}, 2011, pp. 2097--2104.

\bibitem{zhang2016biased}
X.~Zhang, L.~P. Dorado-Munoz, D.~W. Messinger, and N.~D. Cahill, ``Biased
  normalized cuts for target detection in hyperspectral imagery,'' in
  \emph{Algorithms and Technologies for Multispectral, Hyperspectral, and
  Ultraspectral Imagery XXII}, vol. 9840.\hskip 1em plus 0.5em minus
  0.4em\relax SPIE, 2016, pp. 230--240.

\bibitem{gillis2012hyperspectral}
D.~B. Gillis and J.~H. Bowles, ``Hyperspectral image segmentation using
  spatial-spectral graphs,'' in \emph{Algorithms and Technologies for
  Multispectral, Hyperspectral, and Ultraspectral Imagery XVIII}, vol.
  8390.\hskip 1em plus 0.5em minus 0.4em\relax SPIE, 2012, pp. 527--537.

\bibitem{tu2020hyperspectral}
B.~Tu, X.~Yang, N.~Li, C.~Zhou, and D.~He, ``Hyperspectral anomaly detection
  via density peak clustering,'' \emph{Pattern Recognit. Lett.}, vol. 129, pp.
  144--149, 2020.

\bibitem{wang2020mcdpc}
Y.~Wang, D.~Wang, X.~Zhang, W.~Pang, C.~Miao, A.-H. Tan, and Y.~Zhou,
  ``{McDPC}: multi-center density peak clustering,'' \emph{Neural. Comput.
  Appl.}, vol.~32, no.~17, pp. 13\,465--13\,478, 2020.

\bibitem{xu2019generalized}
C.~Xu, ``Generalized {Lasso} problem with equality and inequality constraints
  using {ADMM},'' 2019.

\bibitem{gaines2018algorithms}
B.~R. Gaines, J.~Kim, and H.~Zhou, ``Algorithms for fitting the constrained
  lasso,'' \emph{J Comput Graph Stat.}, vol.~27, no.~4, pp. 861--871, 2018.

\bibitem{boyd2011distributed}
S.~Boyd, N.~Parikh, and E.~Chu, \emph{Distributed optimization and statistical
  learning via the alternating direction method of multipliers}.\hskip 1em plus
  0.5em minus 0.4em\relax Now Publishers Inc, 2011.

\bibitem{chen2012hyperspectral}
Y.~Chen, N.~M. Nasrabadi, and T.~D. Tran, ``Hyperspectral image classification
  via kernel sparse representation,'' \emph{IEEE Trans. Geos. Remote Sens.},
  vol.~51, no.~1, pp. 217--231, 2012.

\bibitem{dong2015maximum}
Y.~Dong, L.~Zhang, L.~Zhang, and B.~Du, ``Maximum margin metric learning based
  target detection for hyperspectral images,'' \emph{ISPRS J. Photogramm.
  Remote Sens.}, vol. 108, pp. 138--150, 2015.

\bibitem{chang2004estimation}
C.-I. Chang and Q.~Du, ``Estimation of number of spectrally distinct signal
  sources in hyperspectral imagery,'' \emph{IEEE Trans. Geos. Remote Sens.},
  vol.~42, no.~3, pp. 608--619, 2004.

\bibitem{7430258}
N.~Acito, S.~Matteoli, A.~Rossi, M.~Diani, and G.~Corsini, ``Hyperspectral
  airborne “viareggio 2013 trial” data collection for detection algorithm
  assessment,'' \emph{IEEE J. Sel. Top. Appl. Earth Obs. Remote Sens.}, vol.~9,
  no.~6, pp. 2365--2376, 2016.

\bibitem{chang2020subspace}
S.~Chang, B.~Du, and L.~Zhang, ``A subspace selection-based discriminative
  forest method for hyperspectral anomaly detection,'' \emph{IEEE Trans. Geos.
  Remote Sens.}, vol.~58, no.~6, pp. 4033--4046, 2020.

\bibitem{huyan2018hyperspectral}
N.~Huyan, X.~Zhang, H.~Zhou, and L.~Jiao, ``Hyperspectral anomaly detection via
  background and potential anomaly dictionaries construction,'' \emph{IEEE
  Trans. Geos. Remote Sens.}, vol.~57, no.~4, pp. 2263--2276, 2018.

\bibitem{fan2021hyperspectral}
G.~Fan, Y.~Ma, X.~Mei, F.~Fan, J.~Huang, and J.~Ma, ``Hyperspectral anomaly
  detection with robust graph autoencoders,'' \emph{IEEE Trans. Geos. Remote
  Sens.}, 2021.

\bibitem{chang2020effective}
C.-I. Chang, ``An effective evaluation tool for hyperspectral target detection:
  3d receiver operating characteristic curve analysis,'' \emph{IEEE Trans.
  Geos. Remote Sens.}, vol.~59, no.~6, pp. 5131--5153, 2020.

\bibitem{wang2021auto}
S.~Wang, X.~Wang, L.~Zhang, and Y.~Zhong, ``Auto-ad: Autonomous hyperspectral
  anomaly detection network based on fully convolutional autoencoder,''
  \emph{IEEE Trans. Geos. Remote Sens.}, vol.~60, pp. 1--14, 2021.

\end{thebibliography}


\end{document}